\documentclass{article}

\usepackage[preprint]{corl_2026} 
\usepackage{graphicx}
\usepackage{booktabs}
\usepackage{amsmath}
\usepackage{amssymb}

\title{Making Foresight Actionable: Repurposing Representation Alignment in World Action Models}

\author{
  Lu Qiu$^{1}$,
  Yizhuo Li$^{\dagger,1}$,
  Yi Chen$^{1}$,
  Yuying Ge$^{2}$,
  Yixiao Ge$^{2}$,
  Xihui Liu$^{\dagger,1}$ \\
  \small{$^{1}$~The University of Hong Kong \quad $^{2}$~XPENG Robotics} \\
  \small{$\dagger$~Corresponding authors} \\
  \small{Project Page: \href{https://xpeng-robotics.github.io/agra}{\texttt{https://xpeng-robotics.github.io/agra}}}
}

\begin{document}
\maketitle
\renewcommand{\thefootnote}{\fnsymbol{footnote}}
\footnotetext[1]{The work was done during the author's internship at XPENG Robotics.} 
\renewcommand{\thefootnote}{\arabic{footnote}} 

\maketitle

\vspace{-20pt}

\begin{figure}[htbp]
    \centering
    \includegraphics[width=1.0\textwidth]{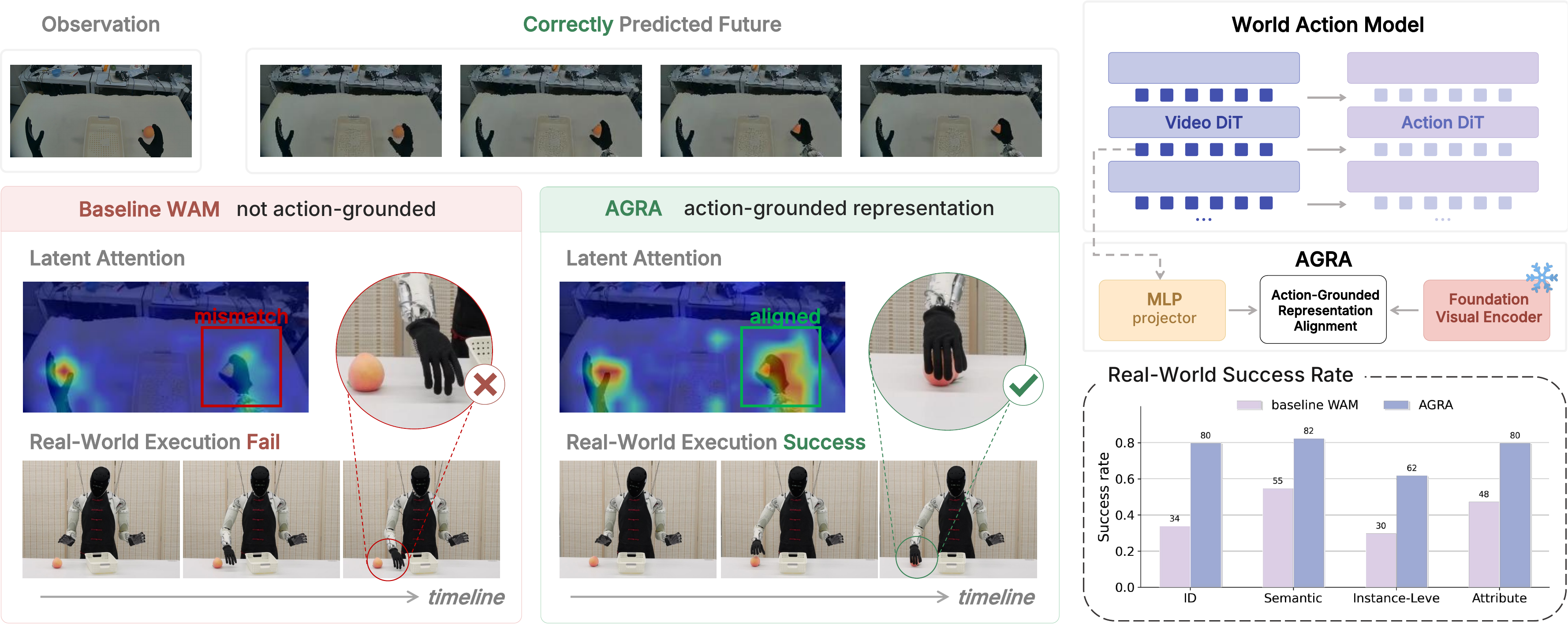}
    \vspace{-10pt}
    \caption{Correctly predicted future in World Action Models does not necessarily yield reliable control, because the visual features can remain poorly organized for action decoding and cause attention to drift toward task-irrelevant regions. We propose AGRA, an Action-Grounded Representation Alignment objective that aligns video features with spatially coherent representations from a frozen foundation visual encoder. This alignment makes world model features more action-grounded and focuses action attention on task-relevant regions, thus leading to higher task success rate.}
    \label{fig:teaser}
\end{figure}

\begin{abstract}
    World Action Models (WAMs) offer a promising route for robot manipulation by using video generation models to model future scene evolution before producing control actions. However, our empirical observations reveal a phenomenon: generating plausible visual futures does not always guarantee the extraction of accurate actions. To diagnose this failure, we conduct action-head attention analysis and causal interventions. We find that the action decoder fails to focus on task-relevant interaction regions and remains sensitive to perturbations in task-irrelevant areas. This reveals a representation mismatch: hidden states optimized for visual reconstruction are not inherently organized in a form useful for low-level action control. In this paper, we propose AGRA, an Action-Grounded Representation Alignment objective that regularizes the world-action interface by aligning intermediate video diffusion features with spatially coherent semantic representations from a foundation visual encoder. We evaluate AGRA on real-world manipulation tasks. Experiments show that AGRA makes world model representations more action-grounded: by focusing the action decoder on the correct interaction regions, it improves object localization accuracy and affordance understanding, and makes the policy more robust to perturbations in task-irrelevant regions. As a result, AGRA consistently improves both in-distribution performance and out-of-distribution generalization over the baseline world action model.  
\end{abstract}



\section{Introduction}

Learning generalist robot policies~\citep{brohan2022rt,zitkovich2023rt,o2024open} requires models to connect visual perception, instructions, and continuous control for robust physical interaction. A natural approach is to equip a policy with predictive model of the world~\citep{cen2025worldvla,zheng2025flare,bi2025motus,team2026motubrain,won2025dual,wu2024unleashing}. World Action Model (WAM)~\citep{ye2026world} first reasons about how scene may evolve and then uses its prediction to guide action generation.~\citep{li2026causal,kim2026cosmos,pai2025mimic} Recent progress in video diffusion models~\citep{brooks2024video,wan2025wan,agarwal2025cosmos} makes this paradigm attractive: large video models can encode rich spatiotemporal structure, predict future, and provide features that contain useful information about motion and task progress in robot manipulation.~\citep{ye2026world,li2026causal,kim2026cosmos,hu2024video}

However, our empirical observations show that plausible visual foresight does not necessarily translate into reliable action prediction. A WAM can generate a plausible and task-consistent future while the action decoder still produces an incorrect motion. To understand this failure mode, we diagnose the world-action interface from both attentional and causal perspectives. We find that the action decoder in the baseline WAM often fails to concentrate on task-relevant interaction regions, such as the hand-object contact areas. In addition, causal interventions on world model hidden states show that action outputs can be sensitive to perturbations in task-irrelevant background regions.

This evidence reveals an \emph{action-grounding gap} between video prediction and action decoding. Video diffusion models are optimized for pixel reconstruction and therefore encode dense appearance-level information, like texture, color, and background clutter~\citep{yu2024representation,zheng2025diffusion,jha2026reconstruction}. However, action prediction does not depend on all visual details in the scene; instead, it is determined by a sparse set of spatially localized and functional factors, such as target objects, contact regions, and affordance geometry. We characterize this mismatch through feature visualization. Compared with foundation visual encoders, video diffusion features are more entangled with low-level appearance. As a result, action decoder is prone to attend to spurious or irrelevant regions, leading to erroneous action predictions.

To close this gap, we introduce \emph{Action-Grounded Representation Alignment} (AGRA). AGRA repurposes representation alignment as a mechanism for action grounding in WAMs. The key idea is to align selected hidden states of the video diffusion model with spatially structured semantic features extracted from a frozen foundation visual encoder. These semantic targets provide a stable reference for the representation field seen by the action decoder: regions with similar semantic or functional roles are encouraged to form coherent structures, while appearance variations become less dominant.

We evaluate AGRA on real-world manipulation tasks. Our analysis shows that by applying AGRA, action decoder attends more accurately to hand-object interaction regions, and becomes more robust to perturbations in task-irrelevant areas. This indicates that AGRA makes the visual representations more action-grounded. By focusing action decoder on the critical interaction regions, AGRA improves object localization accuracy and affordance understanding, while also making the policy more robust to shifts in task-irrelevant regions. As a result, AGRA achieves an in-distribution (ID) success rate of 80\%, substantially outperforming the baseline WAM which obtains 34\%. AGRA also yields stronger robustness under Semantic, Instance-Level, and Attribute Generalization, improving performance by 27\%, 32\%, and 32\%, respectively, compared with the baseline WAM.

Our contributions are threefold. First, we identify and analyze the action-grounding gap in WAMs, showing that plausible visual prediction does not guarantee action-readable representations. Second, we propose AGRA, a representation-alignment objective that regularizes the world-action interface by aligning video model hidden states with spatially coherent pretrained visual features. Third, we validate AGRA in real-world manipulation tasks with the IRON-R01-1.11 humanoid robot, demonstrating improved execution success, stronger out-of-distribution (OOD) generalization, and better cross-embodiment transfer ability.

\section{Related Work}

\textbf{Video Generation as World Modeling.} Recent progress in video generation suggests that predicting future frames is a direct way to learn world dynamics. Large-scale video generators model spatio-temporal structure with autoregressive or diffusion-based objectives, enabling temporally coherent synthesis~\citep{yan2021videogpt,hong2022cogvideo,blattmann2023align,blattmann2023stable,wan2025wan}. This perspective motivates the view that video generators implicitly model scene evolution, object motion, and physical interaction~\citep{brooks2024video,zhu2024sora}, and that's why they are named ``World Models''. Some studies make this connection explicit by developing interactive world models~\citep{bruce2024genie,feng2026matrix,yu2025context,huang2025vid2world}. More recently, embodied world models~\citep{agarwal2025cosmos,chi2025wow,team2025gigaworld,gao2026dreamdojo,shang2026roboscape} have been trained or adapted for physical AI and robot domains. Trained on large-scale robot demonstrations, these models encode richer priors about physical evolution, making them useful for synthetic data generation~\citep{jang2025dreamgen}, evaluation environment~\citep{li2025worldeval,quevedo2025worldgym,guo2025ctrl}, and planning in robotic domains.

\textbf{Video Prediction and Video Representations for Robot Policies.} Video prediction has long served as a predictive interface for robot control. A prominent line of work adopts an explicit generation-and-control paradigm: world models forecast pixel-level future frames under actions~\citep{yu2025context} or instructions, and applying model-predictive control (MPC)~\citep{finn2017deep} or inverse dynamics models (IDMs) to provide policies~\citep{du2023learning, feng2025vidar}. Although intuitive, such explicit methods require full video denoising at inference, slowing down policy execution. Another line of work~\citep{hu2024video,kim2026cosmos,liang2025video,liao2025genie,ma2026dit4dit,pai2025mimic,yuan2026fast} moves to implicit future representations for action prediction. These methods condition policies or IDMs on latent representations of world models that encode plausible future dynamics. More recently, LingbotVA~\citep{li2026causal} employed an autoregressive paradigm for large-scale training, and DreamZero~\citep{ye2026world} jointly modeled video and action, demonstrating strong zero-shot generalization capabilities. However, these world model-based policies directly use predictive features as control inputs, and have not systematically studied whether those representations are suitable for downstream action control.

\textbf{Representation Alignment for Generative Models.} Diffusion Transformers (DiT)~\citep{peebles2023scalable} provide backbone for modern image and video generation. Some studies~\citep{wei2023diffusion,yu2024representation} have shown that the denoising process in generative diffusion
models can induce meaningful representations~\citep{chen2025deconstructing,xiang2023denoising,yang2023diffusion}, but the quality of these representations still lags behind those learned through self-supervised learning methods~\citep{caron2021emerging,oquab2023dinov2,zhai2023sigmoid}. REPA~\citep{yu2024representation} connects these two lines by aligning diffusion model hidden states with representations from foundation visual encoders, improving DiT training efficiency and generation quality. Several variants~\citep{singh2025matters,leng2025repa,hwang2025cross,zhang2026videorepa} extend this idea to enable end-to-end training or improving cross-frame consistency in videos. These methods primarily use representation alignment to improve generative modeling, while AGRA repurposes this for action-grounded robot policy learning. Close to our motivation, a recent study~\citep{jha2026reconstruction} explores VAE encoder choices for world models and finds that semantic spaces better preserve policy-facing structure. But such latent world models~\citep{assran2025v,zhou2024dino} learn dynamics from scratch in compact latent spaces, while WAMs leverage pretrained video representations that already encode physical dynamics from internet-scale data.

\vspace{-5pt}
\section{AGRA: Action-Grounded Representation Alignment}
\label{sec:agra}
\vspace{-3pt}
\subsection{Baseline World Action Model Architecture}
\label{sec:baseline_wam}
The baseline WAM follows a dual DiT architecture, where a world model (Video DiT) generates visual future and an action head (Action DiT) transforms the intermediate future representations into continuous actions. Given the current RGB observation $o_0$, language instruction $c$, and robot proprioceptive state $s_0$, the policy predicts a continuous action chunk with $K$ horizon and $d_a$ dimension:
\begin{equation}
\small
    \hat{\mathbf{a}}_{1:K}
    =
    \pi(o_0, s_0, c),
    \qquad
    \mathbf{a}_{1:K}\in\mathbb{R}^{K\times d_a}.
\end{equation}
Video DiT predicts future frames and exposes its hidden states to action DiT. Let $\mathbf{H}^{\mathrm{vid}}_\ell \in \mathbb{R}^{N_v\times d_v}$ denote the hidden states of video layer $\ell$, for the $j$-th action cross-attention block, a video layer $\ell_j$ is selected uniformly across the depth of the video DiT to use a multi-layer bridge strategy. Let $\mathbf{X}^{\mathrm{act}}$ denote the hidden states of Action DiT, visual representations are injected into action head by:
\begin{equation}
\small
    \mathcal{G}_j
    =
    \mathrm{Proj}_j
    \left(
        \mathbf{H}^{\mathrm{vid}}_{\ell_j}
    \right)
    \in
    \mathbb{R}^{N_v\times d_{\mathrm{act}}}, 
    \mathbf{X}^{\mathrm{act}}_j
    =
    \mathrm{CrossAttn}_j
    \left(
        Q=\mathbf{X}^{\mathrm{act}}_{j-1},
        K=\mathcal{G}_j,
        V=\mathcal{G}_j
    \right).
\end{equation}
Both branches are trained with flow matching~\citep{lipman2022flow}. During action generation, video DiT is evaluated once at a fixed high-noise level $\tau_v^{\mathrm{cond}}=1$ to extract predictive features, and the action DiT performs iterative denoising in action space, keeping inference efficient. In this paper, we use Cosmos-Predict-2.5 as video DiT. For the training objective $\mathcal{L}_{\mathrm{WAM}}$ and more details, please refer to Appendix.

\subsection{Diagnosing the Action-Grounding Gap in World Action Models}
\label{subsec:diagnosis}
A central assumption behind WAMs is that a predictive world representation should provide useful guidance for action decoding. But in real-world deployment, we observe that generating plausible visual futures does not always guarantee the extraction of accurate control actions.

\begin{figure}[t]
\vspace{-5pt}
    \centering
    \includegraphics[width=0.95\textwidth]{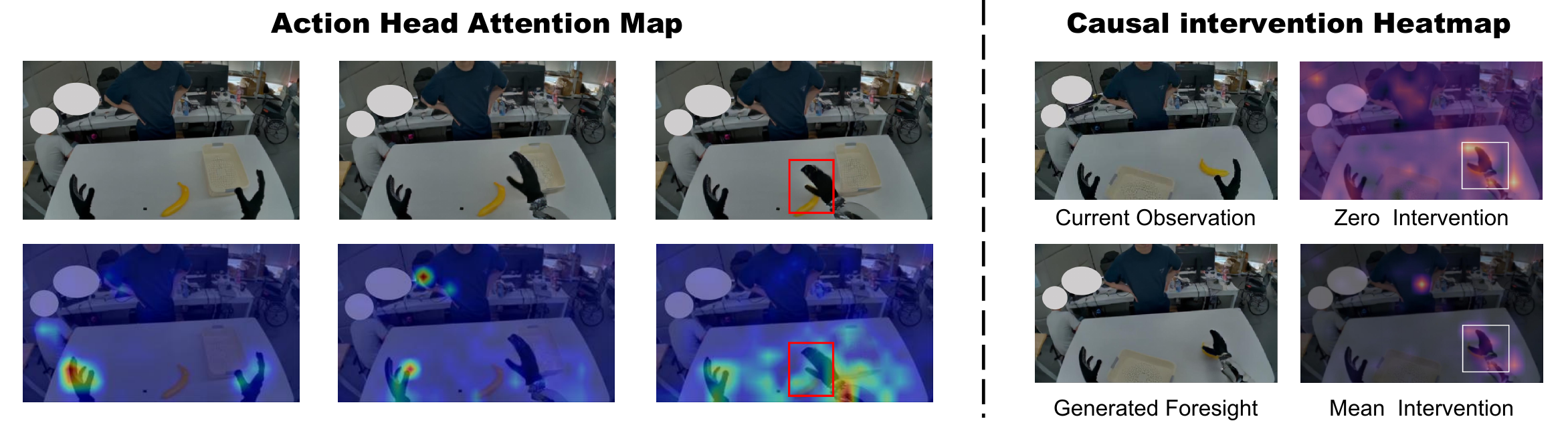}
    \vspace{-8pt}
    \caption{
    \textbf{Diagnosis of the action-grounding gap in the baseline WAM.
(Left)} Cross-attention maps show the action head ignoring the critical hand-object interaction region, despite generating a plausible future video.
\textbf{(Right)} Causal intervention heatmaps (where brighter regions indicate areas that cause the robot’s action to drift when disrupted) reveal that the baseline model’s decisions are heavily influenced by task-irrelevant background elements rather than the interaction area.}
    \label{fig:obser_attn}
    \vspace{-10pt}
\end{figure}

\textbf{Action-head attention analysis.} To diagnose this failure mode, we inspect the cross-attention maps in the action DiT to understand how the action decoder reads from the world model features. We average the attention weights over action tokens and attention heads, and project them back to the video latent shape. In most cases, the attention maps can localize coarse robot hands areas. However, they often fail to concentrate on the most action-critical interaction site. As illustrated in the left panel of Figure~\ref{fig:obser_attn}, the model's attention is distracted by the stationary left hand and the table background, failing to focus on the crucial interaction region between the right hand and the banana.

\textbf{Causal intervention for hidden-state actionability.} Attention maps reveal where the action decoder reads from, but it does not establish which regions causally affect the predicted action. We therefore perform token-level causal intervention on world model hidden states to estimate the actionability of each spatial location. We apply interventions, such as zero or mean value replacement, to each spatial token. We then use Euclidean distance to measure how much each location affects the action, and generate a heatmap by applying min-max normalization. Higher values on the heatmap represent a greater influence on the action. For visualization, we report the map on the last latent frame, where motion and interaction cues are most pronounced. Ideally, the high-impact regions should align with the task-critical hand-object interaction site. However, as demonstrated by an example in the right panel of Figure~\ref{fig:obser_attn}, when applying the Mean Intervention, the region most influencing the action falls on a person in the background. Under the Zero Intervention, the entire irrelevant background exerts a significant impact on the action. This implies that perturbations in the unrelated areas affect the predicted action significantly, making it difficult for the model to maintain robust performance in OOD scenarios, such as altered backgrounds or added distractors.

\begin{figure}[t]
    \centering
    \includegraphics[width=0.95\textwidth]{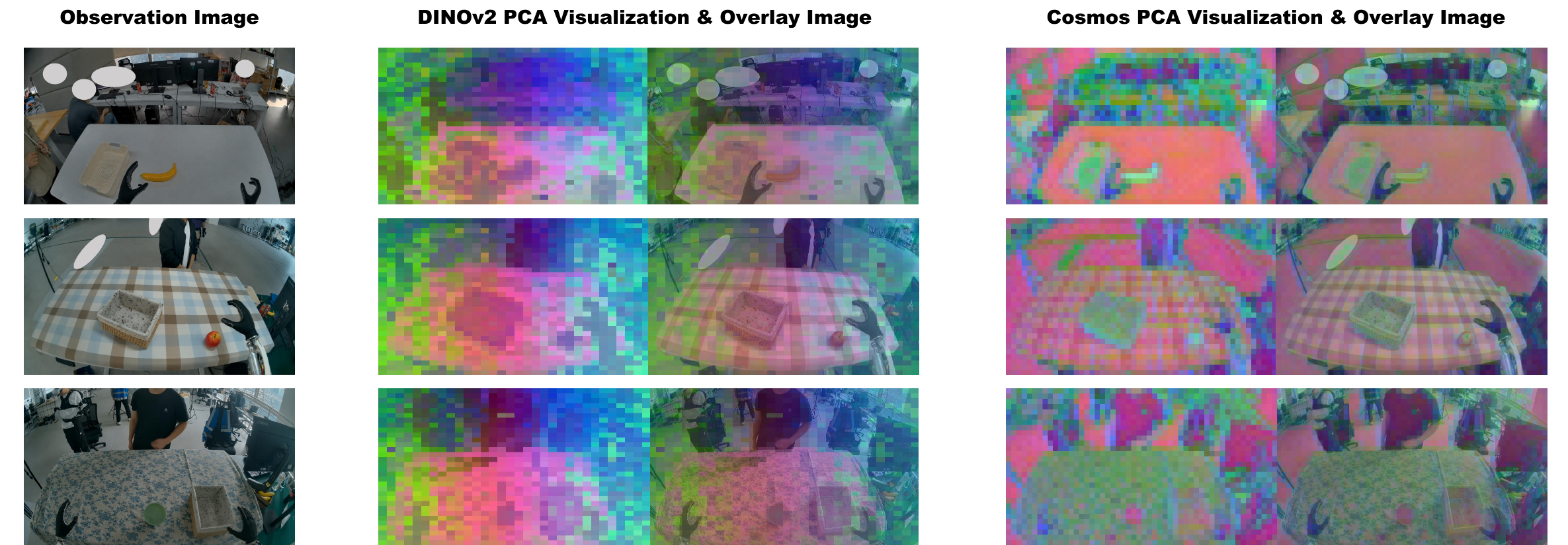}
    \vspace{-8pt}
    \caption{
    PCA visualization of DINOv2 and Cosmos-Predict-2.5 representations. DINOv2 organizes semantically and functionally similar regions into spatially coherent structures, whereas Cosmos features remain more entangled with local appearance.}
    \label{fig:pca_visualize}
    \vspace{-15pt}
\end{figure}

\textbf{Reconstruction-optimized features do not necessarily expose action-readable scene structure.} 
In this paper, we use Cosmos-Predict-2.5 as world model. We visualize the feature space structure of Cosmos and the state-of-the-art self-supervised vision model DINOv2~\citep{oquab2023dinov2} using Principal Component Analysis (PCA)~\citep{abdi2010principal}. For each model, we collect patch features from multiple samples, fit PCA jointly, and map the first three principal components to RGB. As shown in Figure~\ref{fig:pca_visualize}, DINOv2 features exhibit a more coherent spatial organization: semantically and functionally similar regions, such as tables and backgrounds, tend to be mapped to consistent colors across visually different scenes. Even cluttered backgrounds are represented smoothly and separably, and table regions are less sensitive to appearance changes. In contrast, Cosmos features are sensitive to visual details. Background clutter, textured tablecloths, and appearance changes of functionally similar surfaces often produce different features. This suggests that its hidden states expose semantics in a less spatially stable and less accessible form for downstream action decoding. This observation is consistent with the motivation of REPA~\citep{yu2024representation}: although diffusion models can develop meaningful internal representations, their representation quality may still lag behind strong self-supervised visual encoders.

\subsection{Repurposing Representation Alignment for Action Grounding}
\label{subsec:agra_method}

\textbf{Effect of AGRA.} The preceding analysis shows that DINOv2 features provide a more spatially organized and semantically grounded feature space than the native world models. We therefore encourage the guidance visual representations used by the action DiT to align with the DINOv2 feature space. Inspired by REPA, which demonstrates that aligning diffusion hidden states with pretrained visual representations improves their semantic quality, we introduce AGRA: an action-grounded representation alignment objective that regularizes the world-action interface. Unlike the original REPA objective, which is designed to improve diffusion model generation, AGRA regularizes the world-action interface used for robot control.

\textbf{Method.} The first step is to construct DINOv2 targets with the same temporal and spatial shape as world model hidden states. Let the video contain $T_v$ latent frames, with each latent frame covering a set of frames $\{r_t^1,\dots,r_t^q\}$. For the $t$-th latent frame, we use the first corresponding frame $\bar{o}_t=o_{r_t^1}$ as its semantic reference and extract patch-level features from a frozen DINOv2 encoder $g_{\psi}$:
\begin{equation}
\small
    \mathbf{Y}_t
    =
    g_{\psi}(\bar{o}_t)
    \in
    \mathbb{R}^{H_d \times W_d \times d_d}.
\end{equation}
The world model feature is mapped back to a 2D spatial grid $\mathbf{H}^{\mathrm{vid}}\in\mathbb{R}^{T_v \times H_v \times W_v \times d_v}$. To achieve dimensional alignment for DINOv2, we apply spatial interpolation and temporal concatenation:
\begin{equation}
\small
    \tilde{\mathbf{Y}}_t
    =
    \mathrm{Interp}(\mathbf{Y}_t; H_v,W_v)
    \in
    \mathbb{R}^{H_v \times W_v \times d_d}, 
    \tilde{\mathbf{Y}}
    =
    \{\tilde{\mathbf{Y}}_t\}_{t=1}^{T_v}
    \in
    \mathbb{R}^{T_v \times H_v \times W_v \times d_d}.
\end{equation}

The second step is to optimize selected world model hidden states toward these DINOv2 targets. Let $\mathcal{S}_{\mathrm{agra}}=\{\ell_k^{\mathrm{agra}}\}_{k=1}^{K}$ denote the set of world model layers used for alignment. $\mathcal{S}_{\mathrm{agra}}$ can contain either a single layer or multiple layers; in our default setting, unless otherwise specified, we only select one layer for alignment. For each layer, we project the corresponding video hidden state into the DINOv2 feature space via a projector $P_k$:
\begin{equation}
\small
    \mathbf{H}_{\ell_k^{\mathrm{agra}}}^{\mathrm{vid}}
    =
    f_{\theta,\ell_k^{\mathrm{agra}}}^{\mathrm{vid}}
    \left(
        \mathbf{z}^{\tau_v}_{1:T_v},
        o_0,
        c,
        \tau_v
    \right), 
    \mathbf{Z}_k
    =
    P_k
    \left(
        \mathbf{H}_{\ell_k^{\mathrm{agra}}}^{\mathrm{vid}}
    \right)
    \in
    \mathbb{R}^{T_v \times H_v \times W_v \times d_d},
\end{equation}
where $f_{\theta}^{\mathrm{vid}}$ is the world model parameter, $\mathbf{z}^{\tau_v}_{1:T_v}$ is the noisy video latent, and $\tau_v$ is the corresponding noise level. AGRA minimizes the negative cosine similarity between the projected world model hidden states and the aligned DINOv2 features:
\begin{equation}
\small
    \mathcal{L}_{\mathrm{AGRA}}
    =
    -
    \frac{1}{K T_v H_v W_v}
    \sum_{k=1}^{K}
    \sum_{t=1}^{T_v}
    \sum_{u=1}^{H_v}
    \sum_{v=1}^{W_v}
    \cos
    \left(
        \mathbf{Z}_{k,t,u,v},
        \tilde{\mathbf{Y}}_{t,u,v}
    \right), 
    \cos(\mathbf{x},\mathbf{y})
    =
    \frac{
        \mathbf{x}^{\top}\mathbf{y}
    }{
        \|\mathbf{x}\|_2\|\mathbf{y}\|_2+\epsilon
    }.
    \label{eq:agra_loss}
\end{equation}

The final training objective is:
\begin{equation}
    \mathcal{L}
    =
    \mathcal{L}_{\mathrm{WAM}}
    +
    \lambda_{\mathrm{agra}}\mathcal{L}_{\mathrm{AGRA}},
    \label{eq:agra_total_loss}
\end{equation}
where $\mathcal{L}_{\mathrm{WAM}}$ is the training loss of baseline WAM, and $\lambda_{\mathrm{agra}}$ controls the strength of representation alignment. This auxiliary loss does not replace video prediction or action flow matching; it only constrains the intermediate world-model features exposed to the action decoder.

\section{Experiments}

\subsection{Experimental Setup}
We use Cosmos-Predict-2.5-2B with 28 layers as world model, and couple it with an 8-layer action head with 500 M parameters. We evaluate AGRA on the IRON-R01-1.11 humanoid robot, and consider two manipulation tasks, Pick-and-Place and Open-Steamer-Transfer-Bun. The evaluation scenarios includes In-Distribution (ID) setting and three OOD settings (Semantic Generalization, Instance-Level Generalization and Attribute Generalization). More details of the model, dataset, training details, and evaluation setup can be found in Appendix.

\textbf{Compared variants.} We compare following variants. \textit{Freeze backbone} keeps Cosmos frozen and only optimizes action head. \textit{WAM} is the baseline model without representation alignment. \textbf{\textit{AGRA-DinoL8} (ours) aligns the 8th layer of Cosmos with DINOv2 features. This is our default model unless otherwise specified.} \textit{AGRA-DinoL15} aligns a deeper Cosmos layer with DINOv2. \textit{AGRA-DinoL4/8/12} aligns multiple Cosmos layers simultaneously. \textit{AGRA-SiglipL8} replaces DINOv2 with SigLIP as the visual alignment target. \textit{AGRA-BridgeL8} feeds the aligned 8th layer feature repeatedly into all action cross-attention layers. Finally, we compare variants trained with and without EgoDex~\citep{hoque2025egodex} human data to evaluate whether our method improves cross-embodiment transfer.

\subsection{Experimental Results}

\begin{figure}[t]
\vspace{-5pt}
    \centering
    \includegraphics[width=0.96\textwidth]{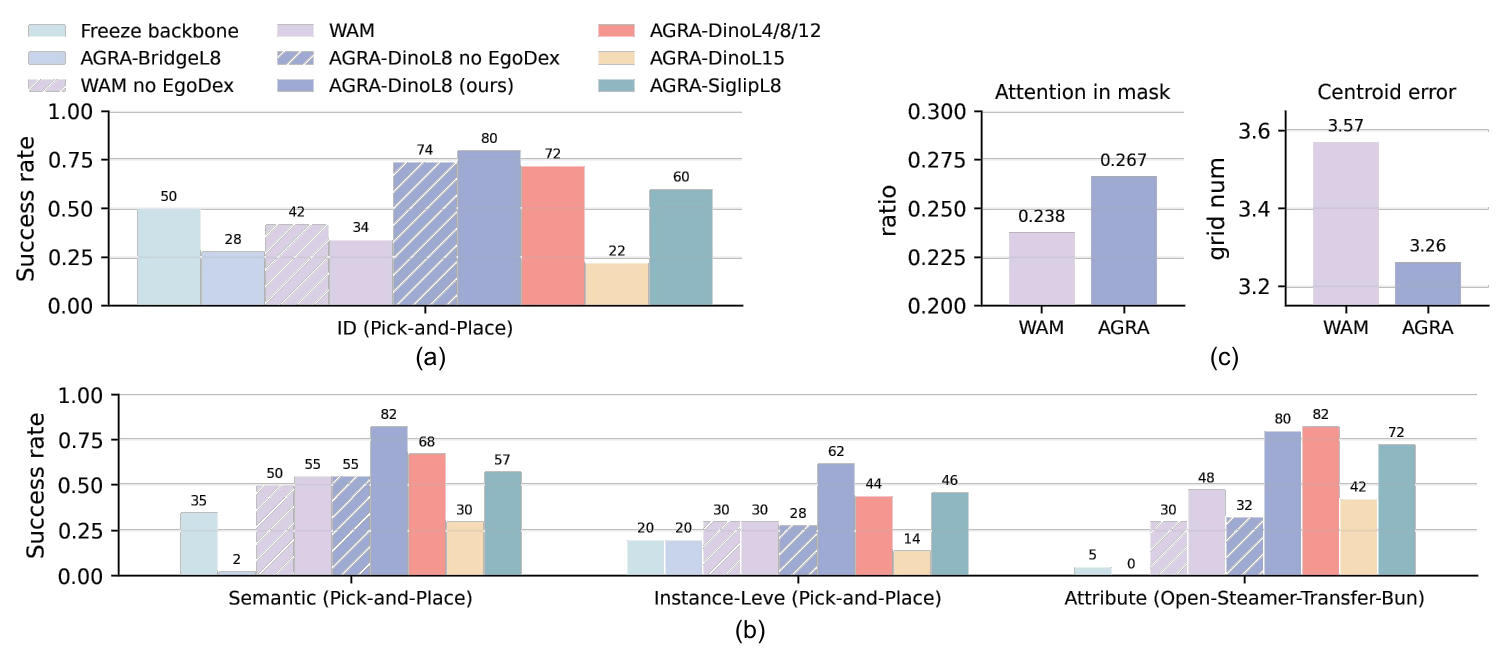}
    \vspace{-10pt}
    \caption{Main results in real-world evaluation. (a) Main results for in-distribution (ID) scenarios. (b) Main results across three out-of-distribution (OOD) generalization scenarios. (c) Quantitative results of the action-head attention analysis.}
    \label{fig:real_world}
    \vspace{-10pt}
\end{figure}

\subsubsection{AGRA Improves Action-Grounded Control}

We first evaluate whether regularizing the world-action interface improves real-world execution. As shown in Figure~\ref{fig:real_world}, the \textit{AGRA} model achieves an ID success rate of 80\%, substantially outperforming the baseline \textit{WAM} which obtains 34\%. \textit{AGRA} also yields stronger robustness under Semantic, Instance-Level, and Attribute Generalization, boosting performance by 27\%, 32\%, and 32\%.

\textbf{Action-head attention analysis.}
\begin{figure}[t]
    \centering
    \includegraphics[width=0.93\textwidth]{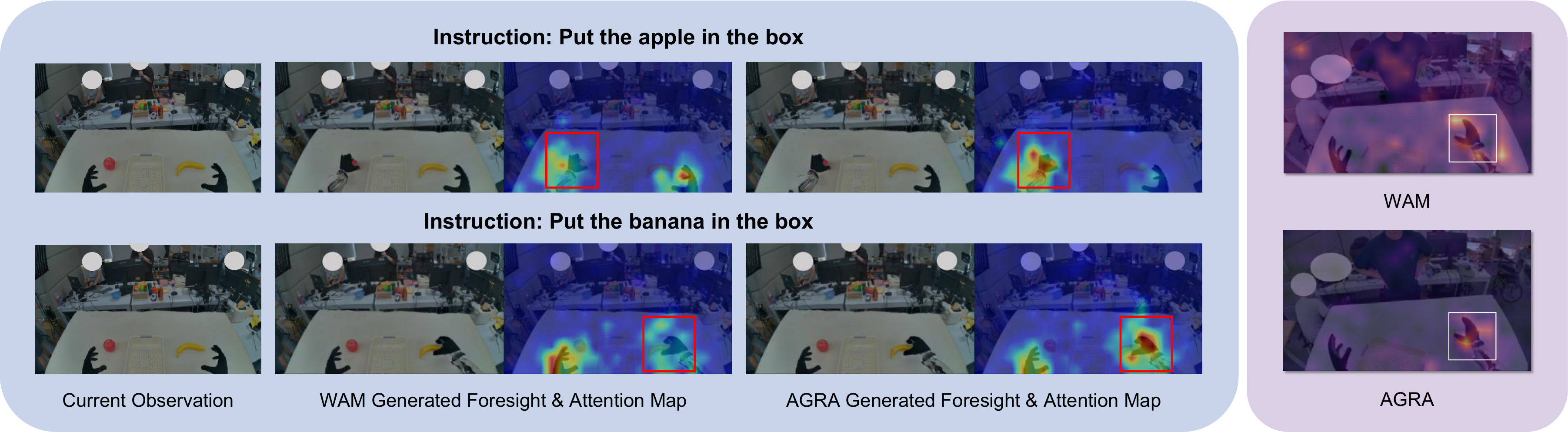}
    \vspace{-5pt}
    \caption{
    Qualitative analysis of action grounding. Compared with baseline WAM, AGRA directs action-head attention and causal sensitivity toward the task-critical hand-object interaction region.}
    \label{fig:attn_map}
    \vspace{-10pt}
\end{figure}
We visualize the cross-attention maps of action DiT in Figure~\ref{fig:attn_map}. We use a Pick-and-Place scene containing an apple and a banana, and evaluate different instructions from the same current observation. Both the baseline \textit{WAM} and \textit{AGRA} can generate plausible future in which the robot arm moves toward the correct object, but they are different in how the action head reads from visual features. For example, in the instruction ``put the banana in the box'', the task-critical region is the interaction area between the right hand and the banana. The \textit{WAM} often allocates large attention mass to regions that are not causally relevant to action, such as the static left hand. In contrast, \textit{AGRA} concentrates more on this task-critical region. We quantify this effect using manually annotated ground-truth hand-object interaction masks on an evaluation subset. We report two metrics: Attention in Mask Ratio, defined as the fraction of total attention mass inside the interaction region, where higher is better; and Centroid Error, defined as the distance between the attention 2D centroid and the center of the annotated region in grid units, where lower is better. Our empirical results shown in Figure~\ref{fig:real_world}(c) indicate that \textit{AGRA} improves both metrics, confirming that \textit{AGRA} makes the action decoder attend more precisely to the spatial regions that determine control.

\begin{figure}[t]
    \centering
    \vspace{-3pt}
    \begin{minipage}[t]{0.44\textwidth}
        \vspace{0pt}
        \centering
        \includegraphics[width=\textwidth]{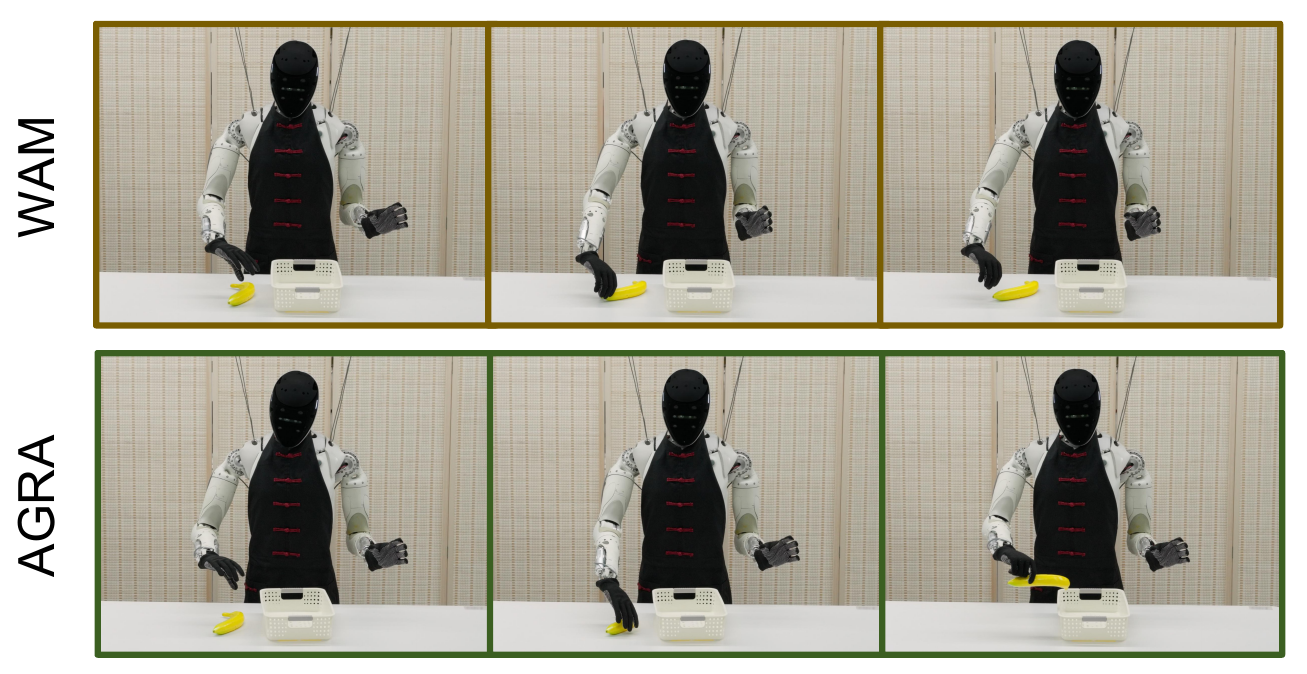}
        \vspace{-20pt}
        \caption{Real-world execution cases.}
        \label{fig:real_world_case}
    \end{minipage}
    \hfill
    \begin{minipage}[t]{0.5\textwidth} 
        \vspace{-3pt}
        \centering
        \makeatletter\def\@captype{table}\makeatother
        \caption{Matched Sensitivity Ratio under various causal intervention strategies.}
        \label{tab:causal_intervention}
        \vspace{7pt}
        \small
        
        \renewcommand{\arraystretch}{1.1} 
        \setlength{\tabcolsep}{10pt}       
        
        \begin{tabular}{ccc} 
        \toprule
        \textbf{Intervention} & \textbf{WAM} & \textbf{AGRA} \\
        \midrule
        Mean & 8.41 & \textbf{10.31} \\
        Zero & 1.95 & \textbf{2.82} \\
        Shuffle & 0.99 & \textbf{1.01} \\
        Swap w/ Comp & 8.36 & \textbf{10.19} \\
        \bottomrule
        \end{tabular}
    \end{minipage}
\vspace{-12pt}
\end{figure}

\textbf{Causal intervention for hidden-state actionability.}
We extend the token-level intervention analysis from Section~\ref{subsec:diagnosis} to obtain a quantitative metric. For each annotated sample, we define the hand-object interaction token set as $\mathcal{I}$, and sample a background token set $\mathcal{B}$ outside this region. We perturb the hidden states for these two sets to obtain the action deviations $\Delta(\mathcal{I})$ and $\Delta(\mathcal{B})$. To eliminate the bias introduced by magnitude, the Matched Sensitivity Ratio is defined as $R ={\Delta(\mathcal{I})}/{\Delta(\mathcal{B})}$. A larger value of $R$ indicates that perturbing the interaction region changes the action more than background region, implying that the hidden states are more action-grounded. We designed different intervention types, including Mean, Zero, Shuffle, and Swap with complement tokens. As shown in Table~\ref{tab:causal_intervention}, \textit{AGRA} consistently achieves a higher $R$ than the baseline. Action-sensitivity heatmaps in Figure~\ref{fig:attn_map} also show the same trend qualitatively: \textit{AGRA} concentrates high-impact regions around the task-critical hand-object contact area, while \textit{WAM} exhibits more diffuse or background-sensitive actionability. This demonstrates \textit{AGRA}'s robustness against task-irrelevant information, explaining its stronger generalization under OOD scenarios (e.g., changed attributes and backgrounds).

\textbf{Real-world execution cases.}
By making visual representations more action-grounded, \textit{AGRA} not only improves generalization by reducing the sensitivity to task-irrelevant noise, but also boosts the basic grasp success rate by focusing the action head on interaction area to enhance object localization and affordance understanding. It translates into more stable physical execution, as shown in Figure~\ref{fig:real_world_case}. The baseline \textit{WAM} often suffers from localization errors, missing the target object due to spatial deviations. \textit{AGRA} produces more stable grasps and better object affordance use. For elongated object such as a banana, \textit{AGRA} accurately targets its center and executes tighter finger closure.

\subsubsection{Further Analysis}
\label{sec:exp_analysis}
\textbf{a. Where Should Representation Alignment Be Applied?}
We next study where representation alignment should be applied inside the world model. As shown in Figure~\ref{fig:real_world}, aligning the 8th layer (\textit{AGRA-DinoL8}) gives better real-world performance than aligning the 15th layer (\textit{AGRA-DinoL15}). This finding suggests that AGRA should be applied at 1/3 of the network depth. By assigning the extraction of semantic representations to shallow layers, the deeper layers are liberated to model motion and fine-grained spatiotemporal dynamics. Enforcing static semantic alignment in deeper layers would disrupt geometric and dynamic details of world model. This conclusion is also well-supported by REPA’s experimental results, and we provide a deeper analysis in Appendix. Furthermore, we compared single-layer with multi-layer aligning. Aligning multiple layers simultaneously, as in \textit{AGRA-DinoL4/8/12}, does not improve performance. We also tested a hierarchical alignment scheme that maps different Cosmos layers to different DINOv2 depths according to their presumed semantic abstraction, but this over-constrains the latent hierarchy and degrades performance.

\textbf{b. Which Visual Representation Is Better for Action Grounding?}
We compare DINOv2 and SigLIP as the alignment target. They provide different representational biases: DINOv2 features are more object-centric and spatially coherent, while SigLIP features are optimized for image-text matching and emphasize global language alignment. 
Therefore, DINOv2 features produce clearer boundaries among objects and background, while SigLIP features are more spatially diffuse and less effective at separating the exact regions involved in manipulation. This can explain why \textit{AGRA-DinoL8} achieves better grasping accuracy and execution stability. Theoretically, SigLIP's language-aligned latent space should bolster the policy's instruction-following capabilities. However, our experiments show that \textit{AGRA-SiglipL8} does not improve semantic generalization. We observe that this is not because \textit{AGRA-SiglipL8} fails to comprehend the language instructions; instead, the performance bottleneck stems from execution-level failures, failing to execute a stable grasp.

\textbf{c. Semantic Information Alone Is Not Sufficient.}
The \textit{AGRA-BridgeL8} variant tests whether an aligned semantic representation alone can drive precise control. In this variant, alignment is applied to 8th layer of Cosmos, and this single-layer feature is repeatedly used as the guidance input for all cross-attention layers of action DiT. This removes the multi-level predictive bridge and isolates the contribution of the aligned semantic layer. As shown in Figure~\ref{fig:real_world}, \textit{AGRA-BridgeL8} performs poorly and drops to 0\% success on the challenging Open-Steamer-Transfer-Bun task. The aligned 8th layer provides strong object identity and scene-layout information, but it lacks the spatial, geometric and dynamic details required for precise manipulation. This confirms that AGRA should be understood as an interface regularizer rather than a replacement for predictive world features.

\textbf{d. Cross-Embodiment Generalization via Human Data.}
Finally, we evaluate whether \textit{AGRA} improves the use of cross-embodiment human data from EgoDex. Human data provides broad visual and interactive diversity, but this diversity is only useful if the policy can extract manipulation-relevant structure that transfers across embodiment. Without action-grounded alignment, the hidden states of a generative model may entangle task structure with embodiment-specific appearance. Empirically, adding EgoDex data to the baseline \textit{WAM} yields little improvement, while \textit{AGRA} benefits substantially from human data, especially in OOD scenarios. 
By anchoring Cosmos hidden states to visual features with stronger semantic coherence, \textit{AGRA} reduces the dependence on embodiment-specific cues and exposes more invariant object-centric and interaction-relevant structure.

\section{Conclusion and Limitation}
In this work, we studied the action-grounding gap between visual representation and action decoding in WAMs. Through attention analysis and causal interventions, we showed that action decoder may attend to task-irrelevant regions and remain sensitive to background perturbations. To address this, we proposed AGRA that regularizes world-action interface by aligning video diffusion features with spatially coherent semantic representations. Experiments on real-world manipulation demonstrate that AGRA leads to more accurate attention over hand-object interaction regions, stronger causal sensitivity to task-relevant areas, improved object localization, and more robust generalization.

\textbf{Limitation.} 
First, our real-world experiments are conducted on a limited set of tasks. Evaluation on more diverse and complex tasks would allow a more comprehensive analysis of different visual representations. For example, we did not observe clear gains from SigLIP, partly because the baseline model performs well on current Pick-and-Place tasks; tasks requiring richer semantic reasoning may better reveal the potential advantages. Second, although we identify the action-grounding problem and provide metrics and methods that can partially mitigate this phenomenon, it does not fully solve the video-action mismatch. Quantitatively characterizing this mismatch remains challenging. We leave broader task coverage and more systematic exploration as future work.



\clearpage


\bibliography{example}  

@article{guo2025ctrl,
  title={Ctrl-world: A controllable generative world model for robot manipulation},
  author={Guo, Yanjiang and Shi, Lucy Xiaoyang and Chen, Jianyu and Finn, Chelsea},
  journal={arXiv preprint arXiv:2510.10125},
  year={2025}
}

@article{huang2025vid2world,
  title={Vid2world: Crafting video diffusion models to interactive world models},
  author={Huang, Siqiao and Wu, Jialong and Zhou, Qixing and Miao, Shangchen and Long, Mingsheng},
  journal={arXiv preprint arXiv:2505.14357},
  year={2025}
}

@article{quevedo2025worldgym,
  title={WorldGym: World Model as An Environment for Policy Evaluation},
  author={Quevedo, Julian and Sharma, Ansh Kumar and Sun, Yixiang and Suryavanshi, Varad and Liang, Percy and Yang, Sherry},
  journal={arXiv preprint arXiv:2506.00613},
  year={2025}
}

@inproceedings{wu2024unleashing,
  title={Unleashing large-scale video generative pre-training for visual robot manipulation},
  author={Wu, Hongtao and Jing, Ya and Cheang, Chilam and Chen, Guangzeng and Xu, Jiafeng and Li, Xinghang and Liu, Minghuan and Li, Hang and Kong, Tao},
  booktitle={International Conference on Learning Representations},
  volume={2024},
  pages={10641--10662},
  year={2024}
}

@article{won2025dual,
  title={Dual-stream diffusion for world-model augmented vision-language-action model},
  author={Won, John and Lee, Kyungmin and Jang, Huiwon and Kim, Dongyoung and Shin, Jinwoo},
  journal={arXiv preprint arXiv:2510.27607},
  year={2025}
}

@article{team2026motubrain,
  title={MotuBrain: An Advanced World Action Model for Robot Control},
  author={Team, MotuBrain and Xiang, Chendong and Bao, Fan and Liu, Haitian and Tan, Hengkai and Bi, Hongzhe and Li, James and Liu, Jiabao and Pang, Jingrui and Jing, Kiro and others},
  journal={arXiv preprint arXiv:2604.27792},
  year={2026}
}

@article{zhou2024dino,
  title={Dino-wm: World models on pre-trained visual features enable zero-shot planning},
  author={Zhou, Gaoyue and Pan, Hengkai and LeCun, Yann and Pinto, Lerrel},
  journal={arXiv preprint arXiv:2411.04983},
  year={2024}
}

@article{li2025worldeval,
  title={Worldeval: World model as real-world robot policies evaluator},
  author={Li, Yaxuan and Zhu, Yichen and Wen, Junjie and Shen, Chaomin and Xu, Yi},
  journal={arXiv preprint arXiv:2505.19017},
  year={2025}
}

@article{shang2026roboscape,
  title={Roboscape: Physics-informed embodied world model},
  author={Shang, Yu and Zhang, Xin and Tang, Yinzhou and Jin, Lei and Gao, Chen and Wu, Wei and Li, Yong},
  journal={Advances in Neural Information Processing Systems},
  volume={38},
  pages={63674--63698},
  year={2026}
}

@article{zhu2024sora,
  title={Is sora a world simulator? a comprehensive survey on general world models and beyond},
  author={Zhu, Zheng and Wang, Xiaofeng and Zhao, Wangbo and Min, Chen and Li, Bohan and Deng, Nianchen and Dou, Min and Wang, Yuqi and Shi, Botian and Wang, Kai and others},
  journal={arXiv preprint arXiv:2405.03520},
  year={2024}
}

@article{jang2025dreamgen,
  title={Dreamgen: Unlocking generalization in robot learning through video world models},
  author={Jang, Joel and Ye, Seonghyeon and Lin, Zongyu and Xiang, Jiannan and Bjorck, Johan and Fang, Yu and Hu, Fengyuan and Huang, Spencer and Kundalia, Kaushil and Lin, Yen-Chen and others},
  journal={arXiv preprint arXiv:2505.12705},
  year={2025}
}

@article{feng2026matrix,
  title={The matrix: Infinite-horizon world generation with real-time moving control},
  author={Feng, Ruili and Zhang, Han and Shu, Zhilei and Yang, Zhantao and Tang, Longxiang and Wang, Zhicai and Zheng, Andy and Xiao, Jie and Liu, Zhiheng and Chu, Ruihang and others},
  journal={Advances in Neural Information Processing Systems},
  volume={38},
  pages={87318--87344},
  year={2026}
}

@article{assran2025v,
  title={V-jepa 2: Self-supervised video models enable understanding, prediction and planning},
  author={Assran, Mido and Bardes, Adrien and Fan, David and Garrido, Quentin and Howes, Russell and Muckley, Matthew and Rizvi, Ammar and Roberts, Claire and Sinha, Koustuv and Zholus, Artem and others},
  journal={arXiv preprint arXiv:2506.09985},
  year={2025}
}

@article{brohan2022rt,
  title={Rt-1: Robotics transformer for real-world control at scale},
  author={Brohan, Anthony and Brown, Noah and Carbajal, Justice and Chebotar, Yevgen and Dabis, Joseph and Finn, Chelsea and Gopalakrishnan, Keerthana and Hausman, Karol and Herzog, Alex and Hsu, Jasmine and others},
  journal={arXiv preprint arXiv:2212.06817},
  year={2022}
}

@inproceedings{zitkovich2023rt,
  title={Rt-2: Vision-language-action models transfer web knowledge to robotic control},
  author={Zitkovich, Brianna and Yu, Tianhe and Xu, Sichun and Xu, Peng and Xiao, Ted and Xia, Fei and Wu, Jialin and Wohlhart, Paul and Welker, Stefan and Wahid, Ayzaan and others},
  booktitle={Conference on Robot Learning},
  pages={2165--2183},
  year={2023},
  organization={PMLR}
}

@inproceedings{o2024open,
  title={Open x-embodiment: Robotic learning datasets and rt-x models: Open x-embodiment collaboration 0},
  author={O’Neill, Abby and Rehman, Abdul and Maddukuri, Abhiram and Gupta, Abhishek and Padalkar, Abhishek and Lee, Abraham and Pooley, Acorn and Gupta, Agrim and Mandlekar, Ajay and Jain, Ajinkya and others},
  booktitle={2024 IEEE International Conference on Robotics and Automation (ICRA)},
  pages={6892--6903},
  year={2024},
  organization={IEEE}
}

@article{cen2025worldvla,
  title={Worldvla: Towards autoregressive action world model},
  author={Cen, Jun and Yu, Chaohui and Yuan, Hangjie and Jiang, Yuming and Huang, Siteng and Guo, Jiayan and Li, Xin and Song, Yibing and Luo, Hao and Wang, Fan and others},
  journal={arXiv preprint arXiv:2506.21539},
  year={2025}
}

@article{zheng2025diffusion,
  title={Diffusion transformers with representation autoencoders},
  author={Zheng, Boyang and Ma, Nanye and Tong, Shengbang and Xie, Saining},
  journal={arXiv preprint arXiv:2510.11690},
  year={2025}
}

@article{jha2026reconstruction,
  title={Reconstruction or Semantics? What Makes a Latent Space Useful for Robotic World Models},
  author={Jha, Saurav and Zholus, Artem and Chandar, Sarath and others},
  journal={arXiv preprint arXiv:2605.06388},
  year={2026}
}

@article{bi2025motus,
  title={Motus: A unified latent action world model},
  author={Bi, Hongzhe and Tan, Hengkai and Xie, Shenghao and Wang, Zeyuan and Huang, Shuhe and Liu, Haitian and Zhao, Ruowen and Feng, Yao and Xiang, Chendong and Rong, Yinze and others},
  journal={arXiv preprint arXiv:2512.13030},
  year={2025}
}

@article{singh2025matters,
  title={What matters for Representation Alignment: Global Information or Spatial Structure?},
  author={Singh, Jaskirat and Leng, Xingjian and Wu, Zongze and Zheng, Liang and Zhang, Richard and Shechtman, Eli and Xie, Saining},
  journal={arXiv preprint arXiv:2512.10794},
  year={2025}
}

@inproceedings{xiang2023denoising,
  title={Denoising diffusion autoencoders are unified self-supervised learners},
  author={Xiang, Weilai and Yang, Hongyu and Huang, Di and Wang, Yunhong},
  booktitle={Proceedings of the IEEE/CVF International Conference on Computer Vision},
  pages={15802--15812},
  year={2023}
}

@inproceedings{yang2023diffusion,
  title={Diffusion model as representation learner},
  author={Yang, Xingyi and Wang, Xinchao},
  booktitle={Proceedings of the IEEE/CVF International Conference on Computer Vision},
  pages={18938--18949},
  year={2023}
}

@inproceedings{chen2025deconstructing,
  title={Deconstructing denoising diffusion models for self-supervised learning},
  author={Chen, Xinlei and Liu, Zhuang and Xie, Saining and He, Kaiming},
  booktitle={International Conference on Learning Representations},
  volume={2025},
  pages={55458--55472},
  year={2025}
}

@inproceedings{wei2023diffusion,
  title={Diffusion models as masked autoencoders},
  author={Wei, Chen and Mangalam, Karttikeya and Huang, Po-Yao and Li, Yanghao and Fan, Haoqi and Xu, Hu and Wang, Huiyu and Xie, Cihang and Yuille, Alan and Feichtenhofer, Christoph},
  booktitle={Proceedings of the IEEE/CVF International Conference on Computer Vision},
  pages={16284--16294},
  year={2023}
}

@article{lipman2022flow,
  title={Flow matching for generative modeling},
  author={Lipman, Yaron and Chen, Ricky TQ and Ben-Hamu, Heli and Nickel, Maximilian and Le, Matt},
  journal={arXiv preprint arXiv:2210.02747},
  year={2022}
}

@article{yuan2026fast,
  title={Fast-WAM: Do World Action Models Need Test-time Future Imagination?},
  author={Yuan, Tianyuan and Dong, Zibin and Liu, Yicheng and Zhao, Hang},
  journal={arXiv preprint arXiv:2603.16666},
  year={2026}
}

@article{hoque2025egodex,
  title={Egodex: Learning dexterous manipulation from large-scale egocentric video},
  author={Hoque, Ryan and Huang, Peide and Yoon, David J and Sivapurapu, Mouli and Zhang, Jian},
  journal={arXiv preprint arXiv:2505.11709},
  year={2025}
}

@article{abdi2010principal,
  title={Principal component analysis},
  author={Abdi, Herv{\'e} and Williams, Lynne J},
  journal={Wiley interdisciplinary reviews: computational statistics},
  volume={2},
  number={4},
  pages={433--459},
  year={2010},
  publisher={Wiley Online Library}
}

@article{zhang2026videorepa,
  title={Videorepa: Learning physics for video generation through relational alignment with foundation models},
  author={Zhang, Xiangdong and Liao, Jiaqi and Zhang, Shaofeng and Meng, Fanqing and Wan, Xiangpeng and Yan, Junchi and Cheng, Yu},
  journal={Advances in Neural Information Processing Systems},
  volume={38},
  pages={122647--122676},
  year={2026}
}

@article{hwang2025cross,
  title={Cross-frame representation alignment for fine-tuning video diffusion models},
  author={Hwang, Sungwon and Jang, Hyojin and Kim, Kinam and Park, Minho and Choo, Jaegul},
  journal={arXiv preprint arXiv:2506.09229},
  year={2025}
}

@inproceedings{leng2025repa,
  title={Repa-e: Unlocking vae for end-to-end tuning of latent diffusion transformers},
  author={Leng, Xingjian and Singh, Jaskirat and Hou, Yunzhong and Xing, Zhenchang and Xie, Saining and Zheng, Liang},
  booktitle={Proceedings of the IEEE/CVF International Conference on Computer Vision},
  pages={18262--18272},
  year={2025}
}

@article{yu2024representation,
  title={Representation alignment for generation: Training diffusion transformers is easier than you think},
  author={Yu, Sihyun and Kwak, Sangkyung and Jang, Huiwon and Jeong, Jongheon and Huang, Jonathan and Shin, Jinwoo and Xie, Saining},
  journal={arXiv preprint arXiv:2410.06940},
  year={2024}
}

@inproceedings{zhai2023sigmoid,
  title={Sigmoid loss for language image pre-training},
  author={Zhai, Xiaohua and Mustafa, Basil and Kolesnikov, Alexander and Beyer, Lucas},
  booktitle={Proceedings of the IEEE/CVF international conference on computer vision},
  pages={11975--11986},
  year={2023}
}

@article{oquab2023dinov2,
  title={Dinov2: Learning robust visual features without supervision},
  author={Oquab, Maxime and Darcet, Timoth{\'e}e and Moutakanni, Th{\'e}o and Vo, Huy and Szafraniec, Marc and Khalidov, Vasil and Fernandez, Pierre and Haziza, Daniel and Massa, Francisco and El-Nouby, Alaaeldin and others},
  journal={arXiv preprint arXiv:2304.07193},
  year={2023}
}

@inproceedings{caron2021emerging,
  title={Emerging properties in self-supervised vision transformers},
  author={Caron, Mathilde and Touvron, Hugo and Misra, Ishan and J{\'e}gou, Herv{\'e} and Mairal, Julien and Bojanowski, Piotr and Joulin, Armand},
  booktitle={Proceedings of the IEEE/CVF international conference on computer vision},
  pages={9650--9660},
  year={2021}
}

@inproceedings{peebles2023scalable,
  title={Scalable diffusion models with transformers},
  author={Peebles, William and Xie, Saining},
  booktitle={Proceedings of the IEEE/CVF international conference on computer vision},
  pages={4195--4205},
  year={2023}
}

@article{ye2026world,
  title={World action models are zero-shot policies},
  author={Ye, Seonghyeon and Ge, Yunhao and Zheng, Kaiyuan and Gao, Shenyuan and Yu, Sihyun and Kurian, George and Indupuru, Suneel and Tan, You Liang and Zhu, Chuning and Xiang, Jiannan and others},
  journal={arXiv preprint arXiv:2602.15922},
  year={2026}
}

@article{li2026causal,
  title={Causal World Modeling for Robot Control},
  author={Li, Lin and Zhang, Qihang and Luo, Yiming and Yang, Shuai and Wang, Ruilin and Han, Fei and Yu, Mingrui and Gao, Zelin and Xue, Nan and Zhu, Xing and others},
  journal={arXiv preprint arXiv:2601.21998},
  year={2026}
}

@article{pai2025mimic,
  title={mimic-video: Video-action models for generalizable robot control beyond vlas},
  author={Pai, Jonas and Achenbach, Liam and Montesinos, Victoriano and Forrai, Benedek and Mees, Oier and Nava, Elvis},
  journal={arXiv preprint arXiv:2512.15692},
  year={2025}
}

@article{liao2025genie,
  title={Genie envisioner: A unified world foundation platform for robotic manipulation},
  author={Liao, Yue and Zhou, Pengfei and Huang, Siyuan and Yang, Donglin and Chen, Shengcong and Jiang, Yuxin and Hu, Yue and Cai, Jingbin and Liu, Si and Luo, Jianlan and others},
  journal={arXiv preprint arXiv:2508.05635},
  year={2025}
}

@article{liang2025video,
  title={Video generators are robot policies},
  author={Liang, Junbang and Tokmakov, Pavel and Liu, Ruoshi and Sudhakar, Sruthi and Shah, Paarth and Ambrus, Rares and Vondrick, Carl},
  journal={arXiv preprint arXiv:2508.00795},
  year={2025}
}

@article{kim2026cosmos,
  title={Cosmos policy: Fine-tuning video models for visuomotor control and planning},
  author={Kim, Moo Jin and Gao, Yihuai and Lin, Tsung-Yi and Lin, Yen-Chen and Ge, Yunhao and Lam, Grace and Liang, Percy and Song, Shuran and Liu, Ming-Yu and Finn, Chelsea and others},
  journal={arXiv preprint arXiv:2601.16163},
  year={2026}
}

@article{feng2025vidar,
  title={Vidar: Embodied video diffusion model for generalist manipulation},
  author={Feng, Yao and Tan, Hengkai and Mao, Xinyi and Xiang, Chendong and Liu, Guodong and Huang, Shuhe and Su, Hang and Zhu, Jun},
  journal={arXiv preprint arXiv:2507.12898},
  year={2025}
}

@article{hu2024video,
  title={Video prediction policy: A generalist robot policy with predictive visual representations},
  author={Hu, Yucheng and Guo, Yanjiang and Wang, Pengchao and Chen, Xiaoyu and Wang, Yen-Jen and Zhang, Jianke and Sreenath, Koushil and Lu, Chaochao and Chen, Jianyu},
  journal={arXiv preprint arXiv:2412.14803},
  year={2024}
}

@article{du2023learning,
  title={Learning universal policies via text-guided video generation},
  author={Du, Yilun and Yang, Sherry and Dai, Bo and Dai, Hanjun and Nachum, Ofir and Tenenbaum, Josh and Schuurmans, Dale and Abbeel, Pieter},
  journal={Advances in neural information processing systems},
  volume={36},
  pages={9156--9172},
  year={2023}
}

@inproceedings{finn2017deep,
  title={Deep visual foresight for planning robot motion},
  author={Finn, Chelsea and Levine, Sergey},
  booktitle={2017 IEEE international conference on robotics and automation (ICRA)},
  pages={2786--2793},
  year={2017},
  organization={IEEE}
}

@inproceedings{yu2025context,
  title={Context as memory: Scene-consistent interactive long video generation with memory retrieval},
  author={Yu, Jiwen and Bai, Jianhong and Qin, Yiran and Liu, Quande and Wang, Xintao and Wan, Pengfei and Zhang, Di and Liu, Xihui},
  booktitle={Proceedings of the SIGGRAPH Asia 2025 Conference Papers},
  pages={1--11},
  year={2025}
}

@article{wan2025wan,
  title={Wan: Open and advanced large-scale video generative models},
  author={Wan, Team and Wang, Ang and Ai, Baole and Wen, Bin and Mao, Chaojie and Xie, Chen-Wei and Chen, Di and Yu, Feiwu and Zhao, Haiming and Yang, Jianxiao and others},
  journal={arXiv preprint arXiv:2503.20314},
  year={2025}
}

@article{team2025gigaworld,
  title={Gigaworld-0: World models as data engine to empower embodied ai},
  author={Team, GigaWorld and Ye, Angen and Wang, Boyuan and Ni, Chaojun and Huang, Guan and Zhao, Guosheng and Li, Haoyun and Zhu, Jiagang and Li, Kerui and Xu, Mengyuan and others},
  journal={arXiv preprint arXiv:2511.19861},
  year={2025}
}

@article{chi2025wow,
  title={Wow: Towards a world omniscient world model through embodied interaction},
  author={Chi, Xiaowei and Jia, Peidong and Fan, Chun-Kai and Ju, Xiaozhu and Mi, Weishi and Zhang, Kevin and Qin, Zhiyuan and Tian, Wanxin and Ge, Kuangzhi and Li, Hao and others},
  journal={arXiv preprint arXiv:2509.22642},
  year={2025}
}

@article{agarwal2025cosmos,
  title={Cosmos world foundation model platform for physical ai},
  author={Agarwal, Niket and Ali, Arslan and Bala, Maciej and Balaji, Yogesh and Barker, Erik and Cai, Tiffany and Chattopadhyay, Prithvijit and Chen, Yongxin and Cui, Yin and Ding, Yifan and others},
  journal={arXiv preprint arXiv:2501.03575},
  year={2025}
}

@article{gao2026dreamdojo,
  title={DreamDojo: A Generalist Robot World Model from Large-Scale Human Videos},
  author={Gao, Shenyuan and Liang, William and Zheng, Kaiyuan and Malik, Ayaan and Ye, Seonghyeon and Yu, Sihyun and Tseng, Wei-Cheng and Dong, Yuzhu and Mo, Kaichun and Lin, Chen-Hsuan and others},
  journal={arXiv preprint arXiv:2602.06949},
  year={2026}
}

@inproceedings{bruce2024genie,
  title={Genie: Generative interactive environments},
  author={Bruce, Jake and Dennis, Michael D and Edwards, Ashley and Parker-Holder, Jack and Shi, Yuge and Hughes, Edward and Lai, Matthew and Mavalankar, Aditi and Steigerwald, Richie and Apps, Chris and others},
  booktitle={Forty-first International Conference on Machine Learning},
  year={2024}
}

@article{brooks2024video,
  title={Video generation models as world simulators},
  author={Brooks, Tim and Peebles, Bill and Holmes, Connor and DePue, Will and Guo, Yufei and Jing, Leo and Schnurr, David and Taylor, Joe and Luhman, Troy and Luhman, Eric and others},
  journal={OpenAI Blog},
  volume={1},
  number={8},
  pages={1},
  year={2024}
}

@article{blattmann2023stable,
  title={Stable video diffusion: Scaling latent video diffusion models to large datasets},
  author={Blattmann, Andreas and Dockhorn, Tim and Kulal, Sumith and Mendelevitch, Daniel and Kilian, Maciej and Lorenz, Dominik and Levi, Yam and English, Zion and Voleti, Vikram and Letts, Adam and others},
  journal={arXiv preprint arXiv:2311.15127},
  year={2023}
}

@inproceedings{blattmann2023align,
  title={Align your latents: High-resolution video synthesis with latent diffusion models},
  author={Blattmann, Andreas and Rombach, Robin and Ling, Huan and Dockhorn, Tim and Kim, Seung Wook and Fidler, Sanja and Kreis, Karsten},
  booktitle={Proceedings of the IEEE/CVF conference on computer vision and pattern recognition},
  pages={22563--22575},
  year={2023}
}

@article{hong2022cogvideo,
  title={Cogvideo: Large-scale pretraining for text-to-video generation via transformers},
  author={Hong, Wenyi and Ding, Ming and Zheng, Wendi and Liu, Xinghan and Tang, Jie},
  journal={arXiv preprint arXiv:2205.15868},
  year={2022}
}

@article{yan2021videogpt,
  title={Videogpt: Video generation using vq-vae and transformers},
  author={Yan, Wilson and Zhang, Yunzhi and Abbeel, Pieter and Srinivas, Aravind},
  journal={arXiv preprint arXiv:2104.10157},
  year={2021}
}

@article{li2025unified,
  title={Unified video action model},
  author={Li, Shuang and Gao, Yihuai and Sadigh, Dorsa and Song, Shuran},
  journal={arXiv preprint arXiv:2503.00200},
  year={2025}
}

@article{chi2025diffusion,
  title={Diffusion policy: Visuomotor policy learning via action diffusion},
  author={Chi, Cheng and Xu, Zhenjia and Feng, Siyuan and Cousineau, Eric and Du, Yilun and Burchfiel, Benjamin and Tedrake, Russ and Song, Shuran},
  journal={The International Journal of Robotics Research},
  volume={44},
  number={10-11},
  pages={1684--1704},
  year={2025},
  publisher={Sage Publications Sage UK: London, England}
}

@article{zheng2025flare,
  title={Flare: Robot learning with implicit world modeling},
  author={Zheng, Ruijie and Wang, Jing and Reed, Scott and Bjorck, Johan and Fang, Yu and Hu, Fengyuan and Jang, Joel and Kundalia, Kaushil and Lin, Zongyu and Magne, Loic and others},
  journal={arXiv preprint arXiv:2505.15659},
  year={2025}
}

@article{ma2026dit4dit,
  title={Dit4dit: Jointly modeling video dynamics and actions for generalizable robot control},
  author={Ma, Teli and Zheng, Jia and Wang, Zifan and Jiang, Chunli and Cui, Andy and Liang, Junwei and Yang, Shuo},
  journal={arXiv preprint arXiv:2603.10448},
  year={2026}
}

@article{lyu2026lda,
  title={Lda-1b: Scaling latent dynamics action model via universal embodied data ingestion},
  author={Lyu, Jiangran and Liu, Kai and Zhang, Xuheng and Liao, Haoran and Feng, Yusen and Zhu, Wenxuan and Shen, Tingrui and Chen, Jiayi and Zhang, Jiazhao and Dong, Yifei and others},
  journal={arXiv preprint arXiv:2602.12215},
  year={2026}
}

@misc{nvidia2025gr00t,
  author = {{NVIDIA GEAR Team} and Azzolini, Allison and Bjorck, Johan and Blukis, Valts and others},
  title = {Gr00t n1.6: An Improved Open Foundation Model for Generalist Humanoid Robots},
  howpublished = {\url{https://research.nvidia.com/labs/gear/gr00t-n1_6/}},
  year         = {2025},
  month        = {December},
}

@article{chen2026dial,
  title={DIAL: Decoupling Intent and Action via Latent World Modeling for End-to-End VLA},
  author={Chen, Yi and Ge, Yuying and Zhou, Hui and Ding, Mingyu and Ge, Yixiao and Liu, Xihui},
  journal={arXiv preprint arXiv:2603.29844},
  year={2026}
}

\newpage
\appendix

\begin{figure}[htbp]
    \centering
    \includegraphics[width=1.0\textwidth]{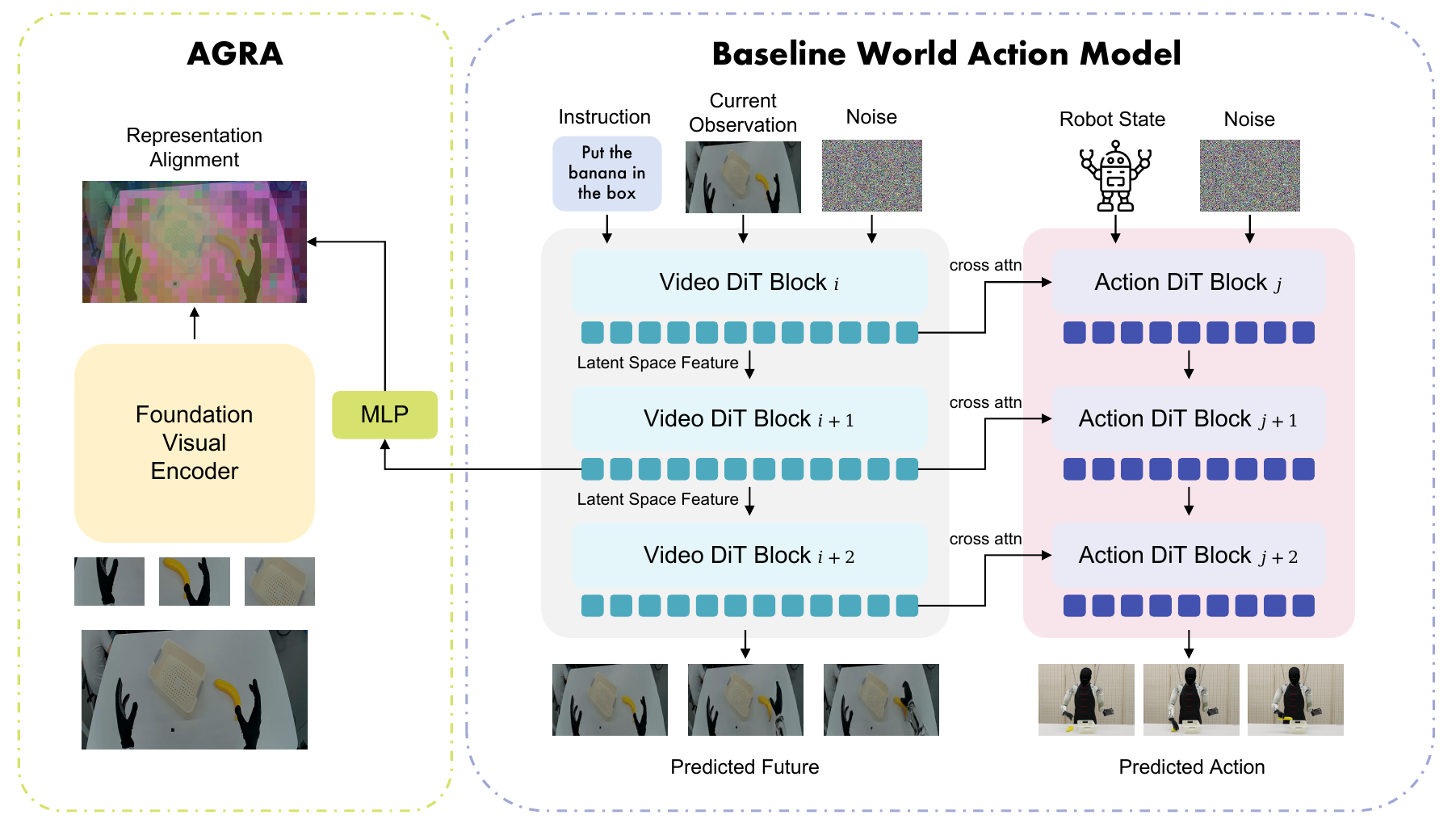}
    \caption{Architecture of baseline World Action Model and the proposed Action-Grounded Representation Alignment.}
    \label{fig:architecture}
\end{figure}

\section{Details of Baseline World Action Model Architecture}
\subsection{Dual-DiT Architecture}
\label{sec:dual_dit_architecture}

\paragraph{Video DiT.} The video branch is a flow-matching video DiT initialized from Cosmos-Predict2.5-2B. It predicts future visual latents conditioned on the current observation $o_0$ and the language instruction $c$. Let $\mathbf{o}_{1:T}$ denote the future video frames and let $\mathbf{z}_{1:T_v} = E_{\mathrm{vae}}(\mathbf{o}_{1:T})$ be the corresponding latent sequence encoded by the video tokenizer. The video DiT predicts the corresponding velocity field at a noise level $\tau_v\in[0,1]$:
\begin{equation}
\begin{gathered}
    \mathbf{z}^{\tau_v}_{1:T_v}
    =
    (1-\tau_v)\mathbf{z}_{1:T_v}
    +
    \tau_v\boldsymbol{\epsilon}_v,
    \qquad
    \boldsymbol{\epsilon}_v\sim\mathcal{N}(\mathbf{0},\mathbf{I}), \\
    \mathbf{v}^{\mathrm{vid}}_\theta
    =
    f^{\mathrm{vid}}_\theta
    (
        \mathbf{z}^{\tau_v}_{1:T_v},
        o_0,
        c,
        \tau_v
    ),
\end{gathered}
\label{eq:video_flow}
\end{equation}

and is optimized with the flow-matching objective:
\begin{equation}
    \mathcal{L}_{\mathrm{vid}}
    =
    \mathbb{E}_{\mathbf{z},\boldsymbol{\epsilon}_v,\tau_v}
    \left[
    \left\|
    f^{\mathrm{vid}}_\theta
    (
        \mathbf{z}^{\tau_v}_{1:T_v},
        o_0,
        c,
        \tau_v
    )
    -
    \left(
        \boldsymbol{\epsilon}_v-\mathbf{z}_{1:T_v}
    \right)
    \right\|_2^2
    \right].
    \label{eq:video_flow_loss}
\end{equation}

The timestep is sampled using the default sampling strategy of Cosmos-Predict2.5-2B. Since the video branch is used as a source of predictive visual representations, we expose its intermediate hidden states to the action branch. We denote the hidden states of the $\ell$-th video DiT layer as:
\begin{equation}
    \mathbf{H}^{\mathrm{vid}}_\ell
    =
    \{h^{\mathrm{vid}}_{\ell,i}\}_{i=1}^{N_v}
    \in
    \mathbb{R}^{N_v\times d_v},
    \label{eq:video_hidden_state}
\end{equation}
where $N_v$ is the number of video latent tokens and $d_v$ is the feature dimension.

\paragraph{Action DiT.}
We employ the flow-matching action DiT from the Gr00T-N1. This action branch
generates a future action chunk from pure noise, conditioned on predictive visual
representations and robot proprioceptive state $s_0$. By sampling the action noise
level $\tau_a$ from a Beta distribution, we construct a noisy action chunk from the
ground-truth action sequence $a_{1:K}$:
\begin{equation}
    a^{\tau_a}_{1:K}
    =
    (1-\tau_a)a_{1:K}
    +
    \tau_a \epsilon_a,
    \qquad
    \epsilon_a \sim \mathcal N(0,I),
    \label{eq:action_noisy_chunk}
\end{equation}
and the noisy action chunk is concatenated with robot proprioceptive state $s_0$
to serve as the action DiT input sequence.

The action DiT alternates self-attention over action / state tokens and
cross-attention to world-model guidance features. At the $j$-th cross-attention
layer, it updates the action / state tokens as:
\begin{equation}
    X^{\mathrm{act}}_j
    =
    \mathrm{CrossAttn}_j
    \left(
        Q=X^{\mathrm{act}}_{j-1},
        K=\mathcal{G}_j,
        V=\mathcal{G}_j
    \right).
    \label{eq:action_cross_attention}
\end{equation}
The action branch predicts action velocity field and is optimized with the
flow-matching objective:
\begin{equation}
\begin{gathered}
    v^{\mathrm{act}}_\phi
    =
    f^{\mathrm{act}}_\phi
    \left(
        a^{\tau_a}_{1:K},
        s_0,
        \tau_a,
        \mathcal{G}
    \right),\\
    \mathcal L_{\mathrm{act}}
    =
    \mathbb E_{a,\epsilon_a,\tau_a}
    \left[
        \left\|
        f^{\mathrm{act}}_\phi
        \left(
            a^{\tau_a}_{1:K},
            s_0,
            \tau_a,
            \mathcal{G}
        \right)
        -
        (\epsilon_a-a_{1:K})
        \right\|_2^2
    \right],
\end{gathered}
\label{eq:action_flow_loss}
\end{equation}
where $\mathcal{G}=\{\mathcal{G}_j\}_{j=1}^{N}$ denotes the set of guidance
features used by the $N$ action cross-attention layers. The concrete construction
of $\mathcal{G}$ during training and inference is described in Sections~\ref{sec:action_sampling}
and~\ref{sec:optimization_strategy}.

\paragraph{Bridge Method.}
This paragraph demonstrates how the hidden states of Video DiT are fed into
Action DiT. A common strategy~\citep{ma2026dit4dit} is to select a single video layer
$\ell^\star$ and reuse its hidden states for all action cross-attention layers,
but it discards the hierarchical structure of the video DiT, which encodes
complementary information at different levels. We therefore use a multi-layer
bridge. Assume the video DiT contains $M$ layers and the action DiT contains
$N$ cross-attention layers. For the $j$-th action cross-attention layer, we select
a video layer uniformly from the depth of the video DiT:
\begin{equation}
    \ell_j
    =
    \left\lfloor
    \frac{j(M-1)}{N-1}
    \right\rceil,
    \qquad
    j=0,\ldots,N-1,
    \label{eq:bridge_layer_selection}
\end{equation}
The selected video hidden states are then projected to the action feature
dimension:
\begin{equation}
    \mathcal{G}_j
    =
    \mathrm{Proj}_j
    \left(
        \mathbf{H}^{\mathrm{vid}}_{\ell_j}
    \right)
    \in
    \mathbb R^{N_v\times d_{\mathrm{act}}},
    \label{eq:bridge_projection}
\end{equation}
and the resulting guidance features are injected into the action DiT through the
cross-attention operation above. This bridge design allows the action head to
access multi-level predictive representations from the video world model while
keeping the action decoder lightweight.

\subsection{Action Sampling}
\label{sec:action_sampling}
Prior video-action policies~\citep{hu2024video,pai2025mimic} have shown that early denoising representations at high noise levels of video diffusion models provide useful
predictive visual features for downstream robotic control. In our preliminary
experiments, action policies conditioned on single-step video denoising features
achieve higher task success than those conditioned on features extracted after
four video denoising steps. This suggests that high-noise video representations
may preserve global task dynamics and future motion cues that are more useful
for action prediction than low-noise representations, which tend to focus more on
visual details.

Therefore, during inference, we run the video DiT once on future latent tokens
at a fixed high-noise level $\tau_v^{\mathrm{cond}}=1$, conditioned on the
current observation $o_0$ and language instruction $c$. We use the layer-wise
hidden-state extractor:
\begin{equation}
    \mathbf{H}^{\mathrm{vid}}_\ell
    =
    f^{\mathrm{vid}}_{\theta,\ell}
    \left(
        \mathbf{z}^{1}_{1:T_v},
        o_0,
        c,
        \tau_v^{\mathrm{cond}}=1
    \right),
    \qquad
    \ell=1,\ldots,M .
    \label{eq:inference_hidden_state}
\end{equation}
The guidance feature for the $j$-th action cross-attention layer is then
computed using the bridge defined in Eq.~\ref{eq:bridge_projection}:
\begin{equation}
    \mathcal{G}_j
    =
    \mathrm{Proj}_j
    \left(
        \mathbf{H}^{\mathrm{vid}}_{\ell_j}
    \right),
    \qquad
    j=1,\ldots,N .
    \label{eq:inference_guidance_feature}
\end{equation}
The lightweight action DiT then performs multi-step denoising in the action
space, conditioned on $\mathcal{G}=\{\mathcal{G}_j\}_{j=1}^{N}$, to generate the
continuous action chunk. This single-step video and multi-step action sampling
strategy reduces inference latency while retaining the predictive guidance
provided by the video world model.

\subsection{Optimization Strategy}
\label{sec:optimization_strategy}

Training WAM involves two flow-matching objectives: the video prediction loss
$\mathcal L_{\mathrm{vid}}$ defined in Eq.~\ref{eq:video_flow_loss} and the action prediction loss
$\mathcal L_{\mathrm{act}}$ defined in Eq.~\ref{eq:action_flow_loss}. The full objective is:
\begin{equation}
    \mathcal L_{\mathrm{WAM}}
    =
    \mathcal L_{\mathrm{vid}}
    +
    \lambda_{\mathrm{act}}\mathcal L_{\mathrm{act}},
    \label{eq:wam_loss}
\end{equation}
where $\lambda_{\mathrm{act}}$ balances the video prediction objective and the
action prediction objective.

For the video generation module, we follow the standard diffusion training paradigm. At each training step, the noise level $\tau_v$ is randomly sampled from [0, 1]. This exposes the video model to all noise levels, forcing it to learn the full denoising trajectory required to synthesize future frames. The loss is computed following Eq.~\ref{eq:video_flow} and Eq.~\ref{eq:video_flow_loss}.

For the action prediction objective, the action head first requires predictive visual
representations from the world model. To keep the visual input to the action
module deterministic and consistent with inference, we compute these
representations using a separate video DiT forward pass at the fixed high-noise
level $\tau_v^{\mathrm{cond}}=1$, which is the same as the inference process in Section~\ref{sec:action_sampling}:
\begin{equation}
    \mathbf{H}^{\mathrm{vid}}_\ell
    =
    f^{\mathrm{vid}}_{\theta,\ell}
    \left(
        \mathbf{z}^{1}_{1:T_v},
        o_0,
        c,
        \tau_v^{\mathrm{cond}}=1
    \right),
    \qquad
    \ell=1,\ldots,M .
    \label{eq:training_hidden_state}
\end{equation}
The multi-layer bridge then constructs the guidance features used by the action
DiT:
\begin{equation}
    \mathcal{G}_j
    =
    \mathrm{Proj}_j
    \left(
        \mathbf{H}^{\mathrm{vid}}_{\ell_j}
    \right),
    \qquad
    \mathcal{G}=\{\mathcal{G}_j\}_{j=1}^{N}.
    \label{eq:training_guidance_feature}
\end{equation}
Conditioned on $\mathcal{G}$, the action DiT is optimized with
$\mathcal L_{\mathrm{act}}$ as defined in Eq.~\ref{eq:action_flow_loss}. By fixing
$\tau_v^{\mathrm{cond}}=1$ for the visual representations used by the action
head, the model trains the action decoder under the same representation
distribution used at inference time.

\section{Simulation Experiment}
\begin{figure}[t]
    \centering
    \includegraphics[width=1.0\textwidth]{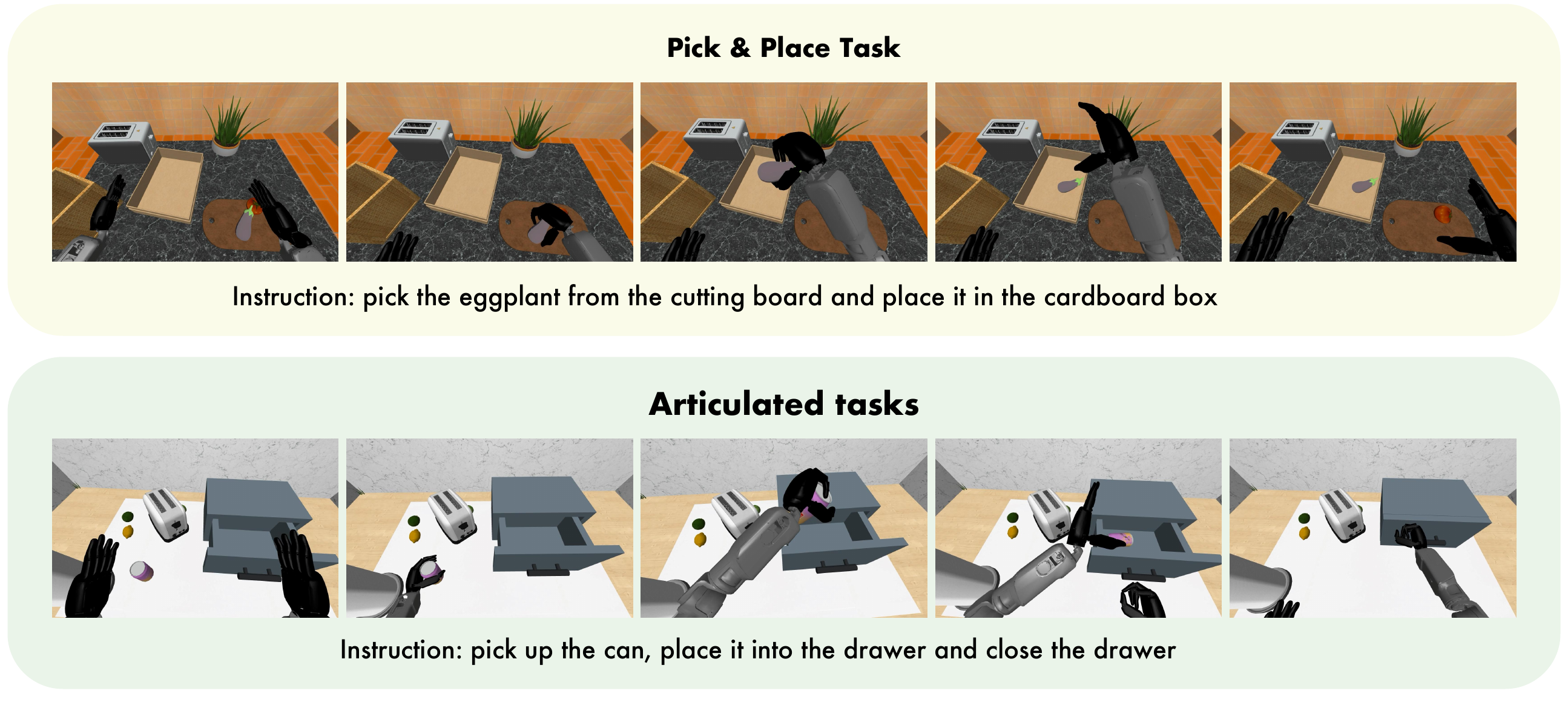}
    \caption{Evaluation results of AGRA on the RoboCasa GR1 benchmark. Videos illustrate representative policy execution trajectories generated by AGRA.}
    \label{fig:gr1_case}
\end{figure}
\begin{figure}[t]
    \centering
    \includegraphics[width=1.0\textwidth]{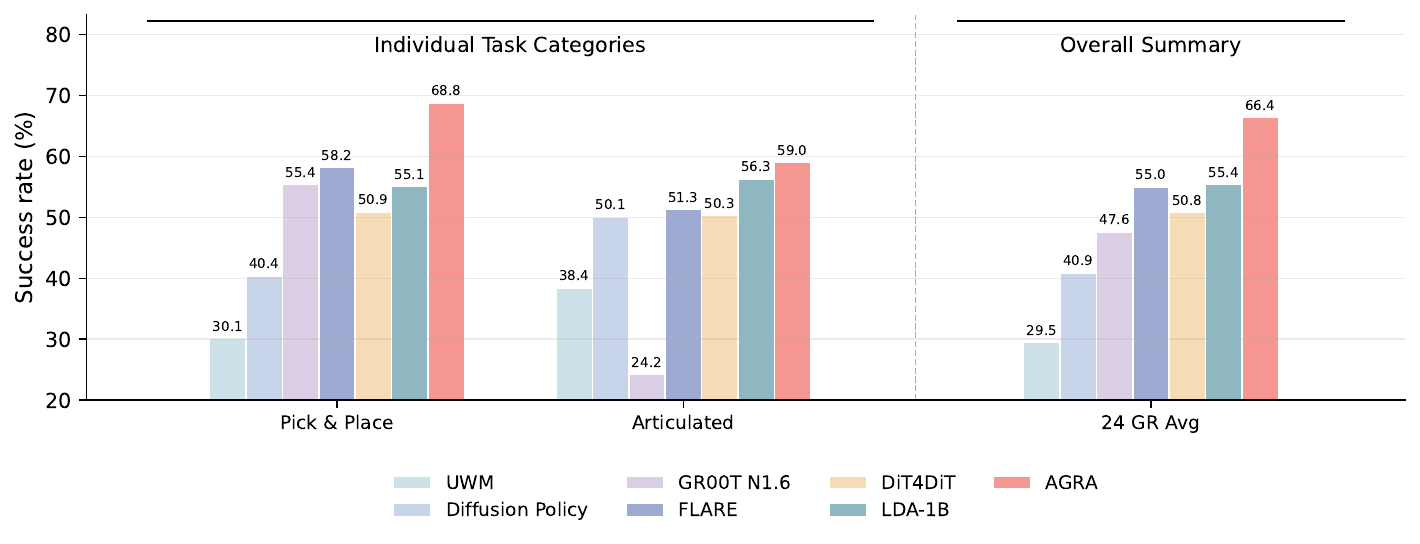}
    \caption{RoboCasa-GR1 tabletop tasks evaluation results with full training data.}
    \label{fig:comparison_methods}
\end{figure}

\begin{figure}[t]
  \centering
  \begin{minipage}[t]{0.48\textwidth} 
    \centering
    \includegraphics[width=\textwidth]{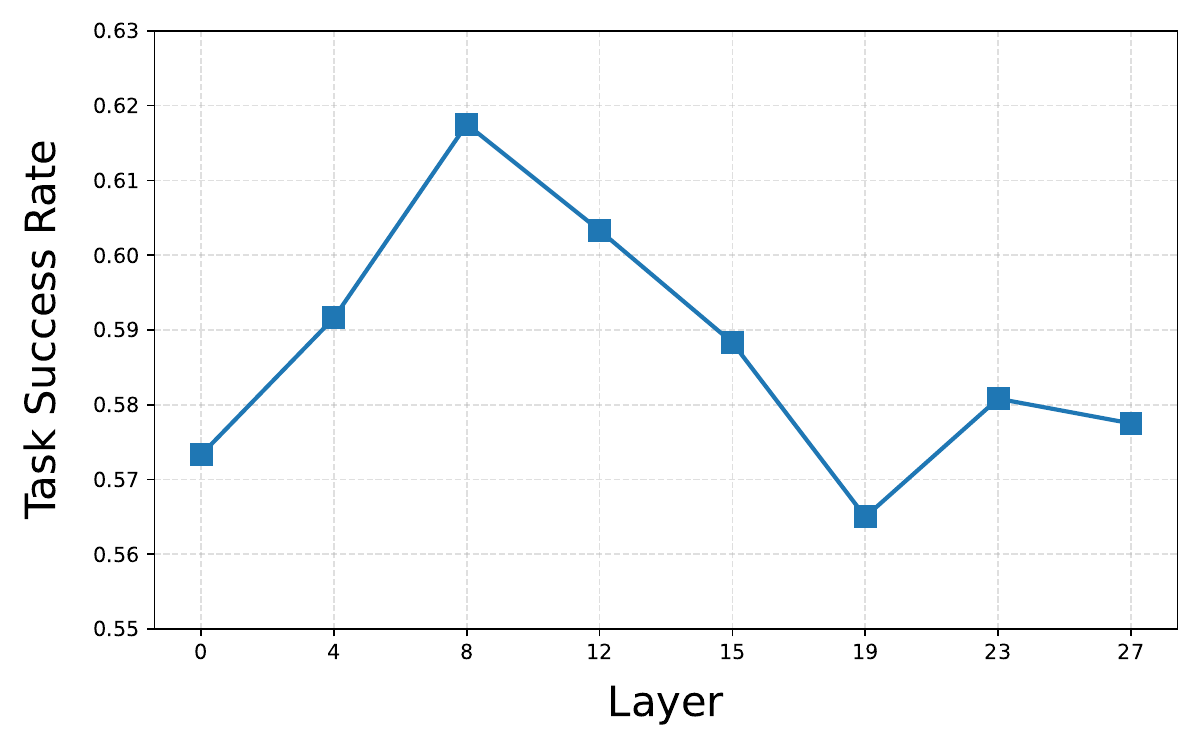}
    \caption{Layer selection for AGRA. The figure visualizes the downstream manipulation performance when applying AGRA to different layers of Cosmos.}
    \label{fig:ablate_layer}
  \end{minipage}
  \hfill
  \begin{minipage}[t]{0.48\textwidth}
    \centering
    \includegraphics[width=\textwidth]{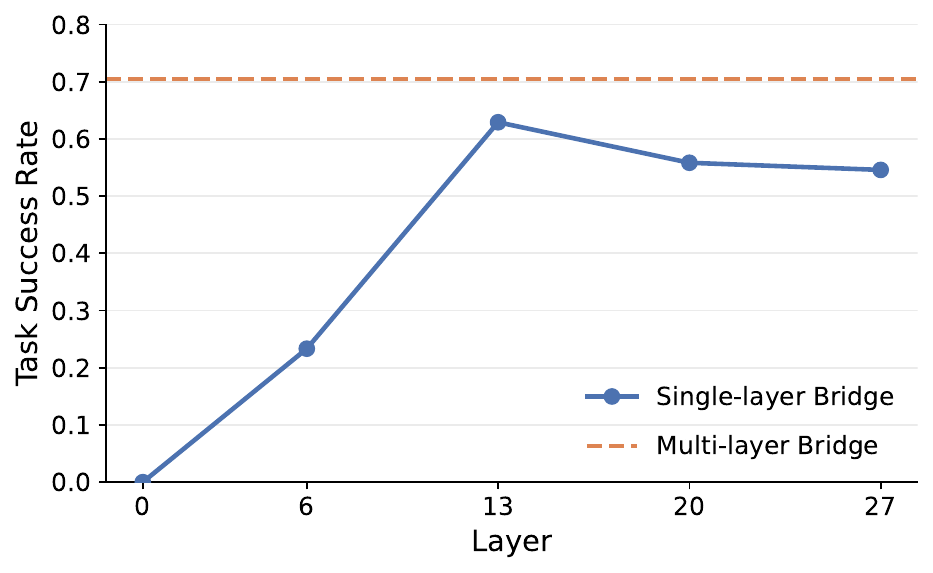}
    \caption{Results of single-layer bridge between dual DiTs. The performance of all single-layer configurations is consistently inferior to that of the multi-layer bridge method.}
    \label{fig:single_layer}
  \end{minipage}
\end{figure}

\begin{table*}[t]
\small
\centering
\renewcommand{\arraystretch}{1.15}
\setlength{\tabcolsep}{8pt}
\caption{Ablation study on the RoboCasa GR1 tabletop benchmark in the few-shot regime.}
\label{tab:robocasa_ablation}
\begin{tabular}{lcccc}
\hline
\textbf{Variant} & \textbf{ID} & \textbf{Unseen App.} & \textbf{Unseen Obj. Types} & \textbf{Unseen Comb.} \\
\hline
Freeze backbone & 52.89 & 49.33 & 43.87 & 51.42 \\
WAM & 58.41 & 53.77 & 43.18 & 59.57 \\
AGRA-SharedDenoisingPass & 58.24 & \textbf{57.77} & 41.87 & 58.57 \\
AGRA-VideoPass & 60.41 & 55.33 & 44.56 & 60.14 \\
AGRA-ActionCondPass & \textbf{61.75} & 56.55 & \textbf{45.31} & \textbf{66.28} \\
\hline
\end{tabular}
\end{table*}

\subsection{RoboCasa GR1 Tabletop Simulation}
We conduct simulation experiments on the RoboCasa benchmark utilizing the GR1 robot. The evaluation suite comprises 24 tabletop tasks, each assessed over 50
episodes. This suite includes 18 ``\textbf{Pick \& Place}'' rearrangement tasks, where the robot follows language instructions to move objects between containers, and 6 ``\textbf{Articulated tasks}'' that involve more complex interactions such as placing objects inside and subsequently closing cabinets, drawers, or microwaves. We represent the robot’s state and actions using 29 joint-space degrees of freedom (DoF), including the dual arms (14), hands (12), and waist (3). For fair comparison with other SOTA methods, we use the full data regime including 24,000 trajectories (1000 per task) and involve 80,000 training steps. For ablation studies, we use few-shot regime with a 10\% subset of 2,400 trajectories (100 per task) with 40,000 training steps, and we also evaluate in three OOD testing scenarios from DIAL~\citep{chen2026dial}: (1) Unseen Appearance (18 tasks), (2) Unseen Object Types (32 tasks), (3) Unseen Combinations (14 tasks).

\subsection{Experimental Results}
\subsubsection{Comparison Against State-Of-The-Art Policies}
We benchmark AGRA against state-of-the-art policies on the comprehensive RoboCasa GR1 Tabletop simulation suite, including UWM~\citep{li2025unified}, a transformer unifying video modeling, forward / inverse dynamics and robot policy; Diffusion Policy~\citep{chi2025diffusion}, which represents robot policy as denoising diffusion process; FLARE~\citep{zheng2025flare}, a flow-matching framework with future latent alignment; DiT4DiT~\citep{ma2026dit4dit}, a video action model coupling video diffusion model with an action head; and LDA-1B~\citep{lyu2026lda}, which learns scalable dynamics in DINO latent space. For VLA, we include GR00T-N1.6~\citep{nvidia2025gr00t} due to its widespread adoption and established efficacy in the robotics community. As illustrated in Figure~\ref{fig:comparison_methods}, AGRA achieves an overall success rate of 66.4\%, outperforming the strong VLA baseline, GR00T, by a significant absolute margin of 18.8\%. When compared with contemporary predictive and generative control methods, including FLARE, DiT4DiT, and LDA-1B—AGRA, consistently demonstrates a performance improvement exceeding 10\%. 

\subsubsection{Ablation study}
To isolate the impact of AGRA's architectural choices, we construct several controlled variants.
\begin{itemize}
    \item \textit{Freeze backbone}: It keeps Cosmos frozen and only optimizes action head.
    \item \textit{WAM}: It is the baseline model without representation alignment.
    \item \textit{AGRA-SharedDenoisingPass}: It removes the separate action-conditioning forward pass. During training, the action head directly consumes the Cosmos hidden states produced by the video denoising pass at the sampled diffusion timestep $\tau_v$, i.e., the same forward pass used to compute $\mathcal{L}_{vid}$.
    \item \textit{AGRA-VideoPass}: It keeps the default action-conditioning branch, but applies AGRA alignment loss to the video denoising pass. The aligned hidden states are taken from the Cosmos forward pass at the sampled video noise level $\tau_v$, the same pass used to compute $\mathcal{L}_{vid}$.
    \item \textit{AGRA-ActionCondPass}: The AGRA alignment loss is applied to the action-conditioning pass, where Cosmos is evaluated at the fixed high-noise level $\tau_v^{cond}=1$, and the resulting hidden states are used by the action head to compute $\mathcal{L}_{act}$.
\end{itemize}

\paragraph{Importance of video-action joint modeling.}
Freezing the world model backbone after pure-video finetuning yields suboptimal performance, establishing a notably low performance ceiling. We observe that under the \textit{Freeze backbone} setting, the action head tends to overfit to proprioceptive states while ignoring the visual signals. This indicates that purely generative pretraining is insufficient for deriving robust control policies. Jointly optimizing the world model and the action head allows the action gradients to backpropagate into the world model, which is critical for morphing the observation-reconstruction hidden states into an action-conditioned representation space.

\paragraph{Effectiveness of AGRA.}
Removing AGRA from the jointly trained model (\textit{WAM}) leads to a deterioration across all metrics, with a particularly steep decline in out-of-distribution generalization, dropping from 66.28 to 59.57 on Unseen Combinations. Without explicit semantic anchoring, the generative latents remain heavily entangled with high-frequency rendering details. By aligning the world model's hidden states with DINOv2 features, AGRA provides a structural regularization that grounds the representations, enhancing the action decoder's ability to generalize across novel object attributes and spatial configurations. However, AGRA shows marginal gains in simulation compared to real-world settings. Failure case analysis indicates that in simpler simulated backgrounds, the baseline model often grasps the object successfully but suffers from post-grasp slippage, which cannot be solved purely through visual improvements.

\paragraph{Layer selection for AGRA.}
As shown in Figure~\ref{fig:ablate_layer}, we also evaluated the impact of applying AGRA constraints at different layers in simulation benchmark. As the application depth increases, the downstream task success rate initially rises before declining, peaking at Layer 8. Consistent with the real-world experiments in Section~\ref{sec:exp_analysis} (a), this reaffirms that applying AGRA at shallow layers, approximately one third of the total depth of the world model, yields optimal performance.

\paragraph{Train-inference consistency in denoising.}
Our ablation on the denoising timestep strategy reveals that using the shared video denoising pass for action conditioning, i.e., \textit{AGRA-SharedDenoisingPass}, yields inferior results compared to the default configuration. By forcing the world model to perform a separate forward pass at pure noise to extract hidden states for the action head in training, the model explicitly enforces train-inference consistency. This design constraint focuses the action head's learning capacity on the high-noise video representations that it will consume at inference time.

\paragraph{Targeted application of AGRA.}
We investigate the specific placement of the AGRA objective. \textit{AGRA-VideoPass}, which applies alignment during the random-timestep forward pass used for calculating the diffusion loss, fails to match the performance of applying AGRA to the pure-noise forward pass dedicated to the action head, i.e., \textit{AGRA-ActionCondPass}. Imposing representation alignment directly on the exact hidden states consumed by the action decoder provides a more explicit and effective semantic grounding, leading to superior empirical robustness.

\paragraph{Multi-layer feature aggregation.}
Using the same setup, we contrasted single- vs. multi-layer bridges between the Video and Action DiTs. Figure~\ref{fig:single_layer} illustrates that although mid-to-deep layers perform best within single-layer setups, they are consistently outperformed by the multi-layer approach. This validates that robotic manipulation requires a hierarchical synthesis of information: multi-layer aggregation is indispensable for simultaneously capturing the high-level object-centric topologies and the fine-grained, low-level spatial dynamics necessary for precise continuous control.

\section{Implementation Details}
\subsection{Model Framework}
We build our policy on top of the Cosmos-Predict-2.5-2B video diffusion backbone. The model takes 17-frame video clips at a resolution of 192 x 336, which are encoded into 5 latent frames using a temporal compression ratio of 4. During training, the first latent frame is used as the visual condition. We extract intermediate visual features from Cosmos transformer layers 0, 4, 8, 12, 15, 19, 23, and 27. These multi-layer features are passed through a project module with layer normalization and a 4-layer self-attention transformer. The action head is a flow-matching diffusion policy head conditioned on the processed Cosmos features. It uses a 8-block DiT with 32 attention heads, hidden size 1024, and cross-attention dimension 2048. Each block contains one self-attention layer and one cross-attention layer. Therefore, the action head has 8 cross-attention layers in total, which are in one-to-one correspondence with the 8 selected Cosmos feature layers. In the default setting, we apply AGRA on Cosmos layer 8. The target representation is extracted from a frozen DINOv2 encoder with input size 448 x 448. At inference time, Cosmos takes the current observation image as the first-frame condition and pure noise for the remaining latent frames. After one denoising step, Cosmos hidden states are extracted and used to condition the action head. The action head then performs 4 flow-matching denoising steps to produce the final action trajectory.

\subsection{Real-World Experimental Setup}
\subsubsection{Dataset Collection}
In real-world experiment, we design two tasks:
\begin{itemize}
    \item \textbf{Pick-and-Place}: The robot must follow the instruction to pick up an object and place it into a container.
    \item \textbf{Open-Steamer-Transfer-Bun}: Expanding on the basic Pick-and-Place operation, this task requires the robot to use its left hand to remove and set aside the steamer lid, and its right hand to pick the bun from inside the steamer and transfer it to a plate on the right. This is a more challenging bimanual cooperative task.
\end{itemize}
For pretraining, we collect a real-world robotic dataset in an industrial factory setting, containing 40k trajectories (92 hours), for the Pick-and-Place task. We additionally use a basic Pick-and-Place subset from EgoDex, which contributes 37k human demonstration trajectories (42 hours). The joint pretraining dataset is used to train the model for 60,000 steps. For laboratory deployment, we collect fine-tuning data: 560 trajectories (1.5 hours) for Pick-and-Place task with a wide variety of objects, and 180 trajectories for Open-Steamer-Transfer-Bun task with 3 types of buns and plates of different colors. The laboratory dataset is used for 2,000 fine-tuning steps.

\begin{figure}[htbp]
    \centering
    \includegraphics[width=1.0\textwidth]{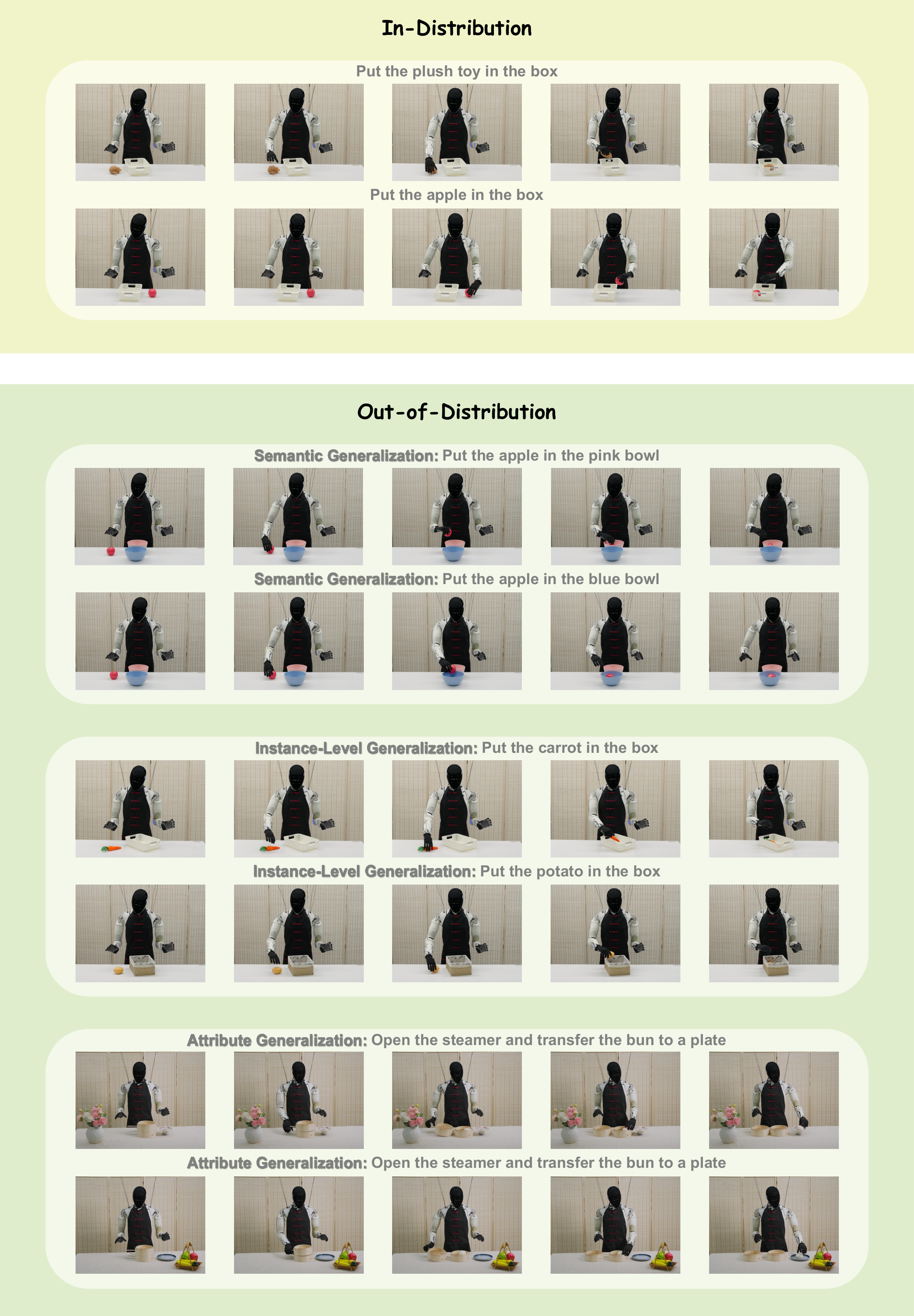}
    \caption{
    Additional real-world execution cases of AGRA. The trajectories show representative AGRA rollouts under different evaluation settings, including in-distribution pick-and-place, semantic generalization, instance-level generalization, and attribute generalization.
    }
    \label{fig:demo}
\end{figure}

\begin{figure}[htbp]
    \centering
    \includegraphics[width=1.0\textwidth]{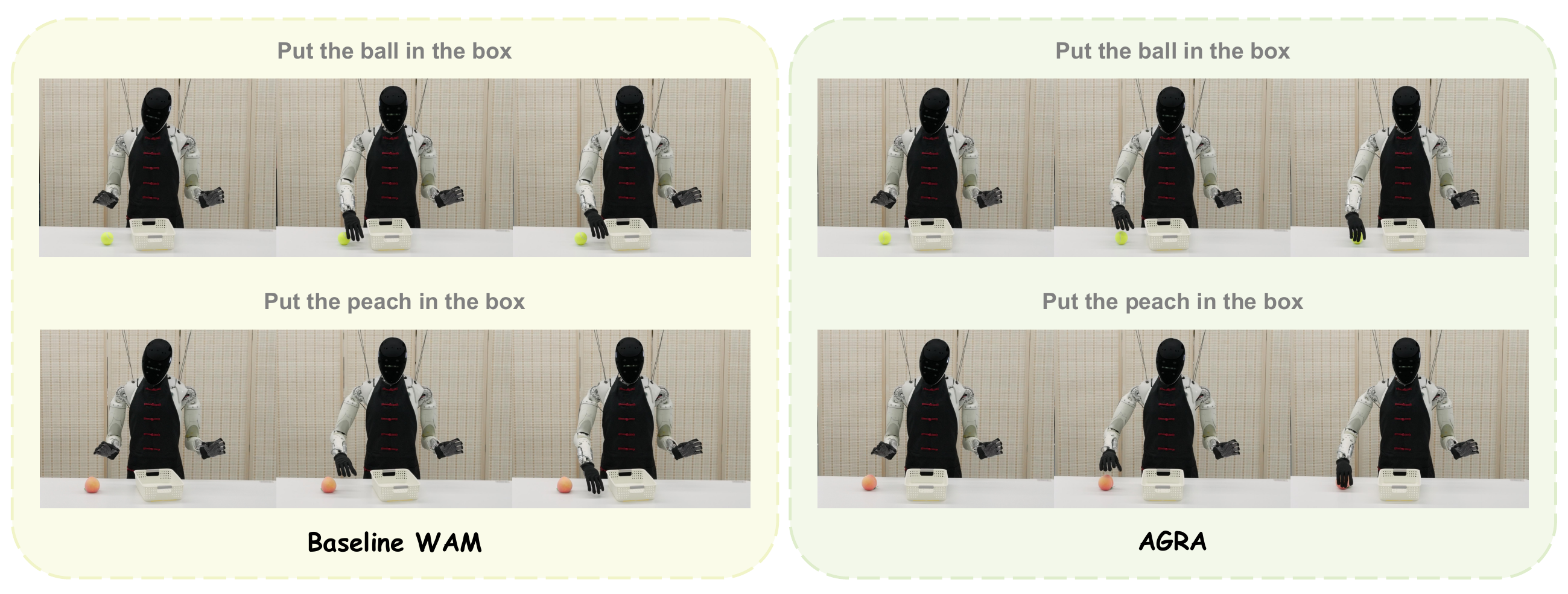}
    \caption{Real-world comparison between WAM and AGRA. Given the instructions ``put the ball in the box'' and ``put the peach in the box'', WAM fails to localize the target accurately, while AGRA reaches and grasps the correct object region.}
    \label{fig:demo_compare}
\end{figure}

\begin{figure}[t]
    \centering
    \includegraphics[width=1.0\textwidth]{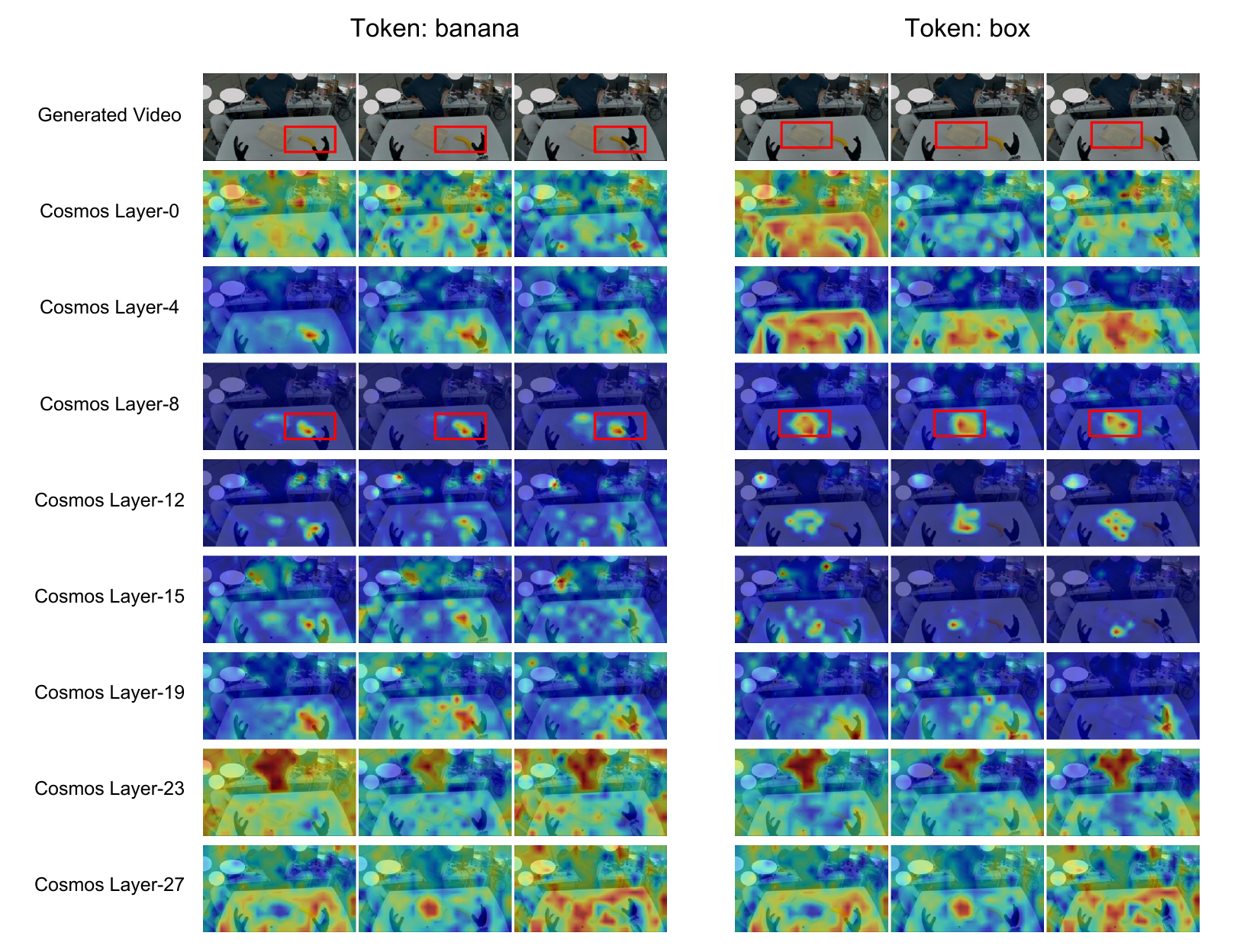}
    \caption{Text-to-video cross-attention visualization across Cosmos layers. We visualize the cross-attention maps for the instruction tokens ``banana'' and ``box'' in the generated video. The 8th layer produces the clearest correspondence between each text token and its associated visual region.}
    \label{fig:all_layer_text}
\end{figure}

\begin{figure}[t]
    \centering
    \includegraphics[width=1.0\textwidth]{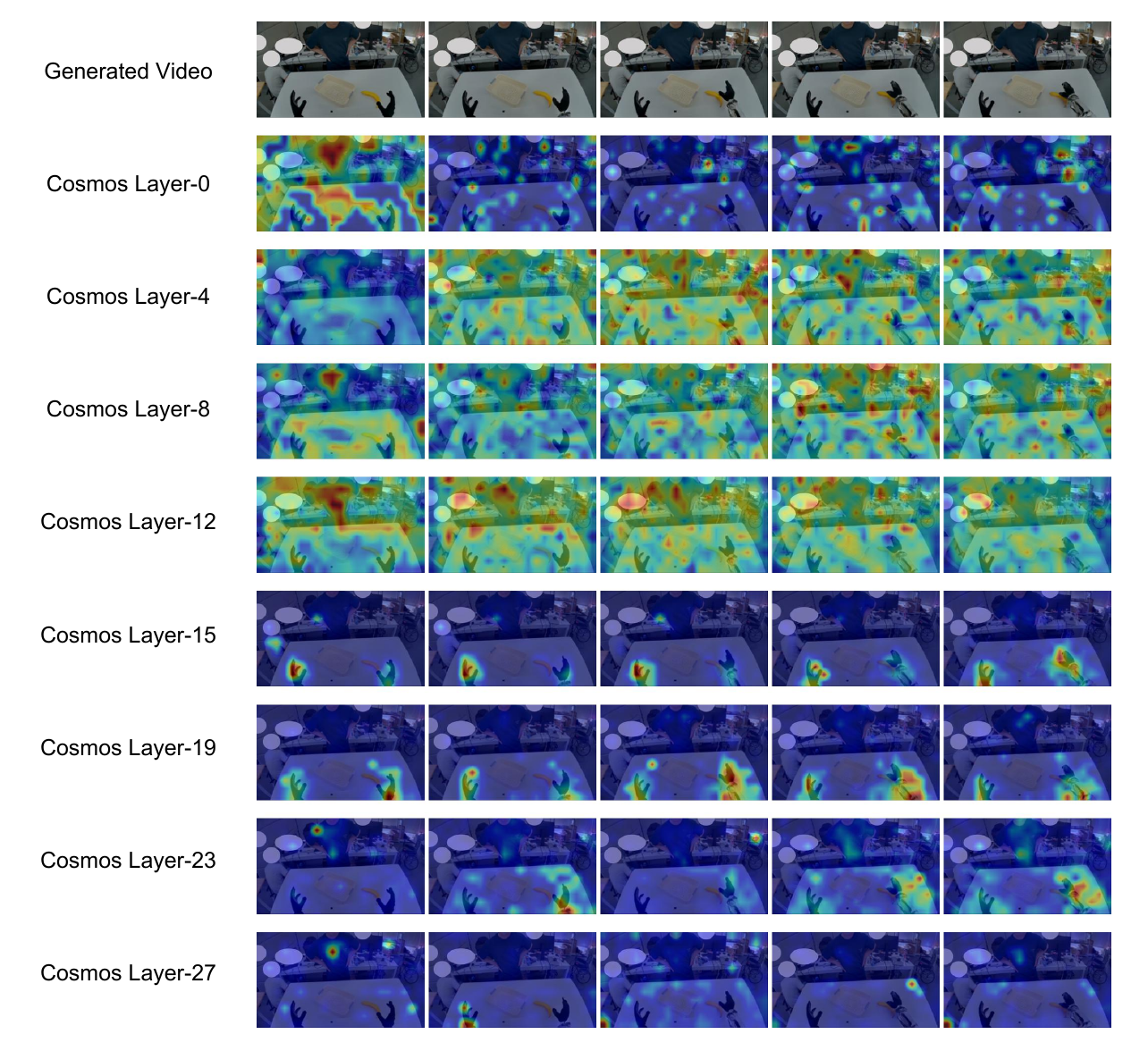}
    \caption{Action-head cross-attention visualization. We visualize where the action head attends when predicting actions from Cosmos video representations. Early layers mainly attend to global scene information, while middle and later layers gradually concentrate on the robot hands and hand-object interaction regions.}
    \label{fig:all_layer_action}
\end{figure}

\begin{figure}[t]
    \centering
    \includegraphics[width=1.0\textwidth]{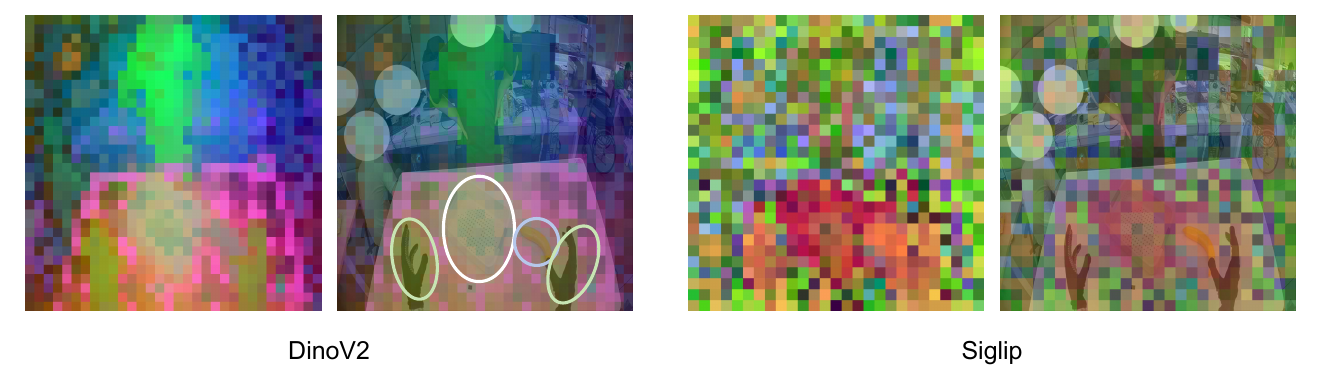}
    \caption{Visualization of different semantic feature types. We compare PCA visualizations of DINOv2 and SigLIP features. DINOv2 exhibits a more object-centric and spatially coherent feature structure, with clearer object boundaries, geometry, and part-level separation.}
    \label{fig:dino_feature}
\end{figure}

\subsubsection{Evaluation}
To comprehensively validate the robustness of the models in physical deployments, we establish an in-distribution (ID) evaluation regime and three out-of-distribution (OOD) generalization regimes. All evaluations are conducted on the real humanoid robot, and each trial is counted as successful only when the robot completes the instructed manipulation goal.
\begin{itemize}
\item \textbf{In-Distribution (ID):} Evaluated on the Pick-and-Place task, this regime measures the policy's ability to manipulate objects that have been seen in training corpus. We select 10 seen objects and place each object at 5 pre-designed tabletop positions that cover different reachable regions and relative spatial configurations. This results in 50 real-world trials.

\item \textbf{Semantic Generalization (OOD):} Evaluated on the Pick-and-Place task, the robot is required to accurately grasp a specified object from several distractors or place it into a designated container following the user's instruction. We design 8 scenarios in which the robot must pick a designated object from several distractors, and 2 scenarios in which the robot must place an object into a specified target container. For each scenario, we specify two candidate targets or containers, denoted as $a$ and $b$. We evaluate both instructions, e.g., picking $a$ and picking $b$, or placing the object into container $a$ and container $b$. We then swap the spatial positions of $a$ and $b$ and repeat the two instructions, which tests whether the policy follows the semantic instruction rather than a fixed spatial bias. Therefore, each scenario contains 4 trials, giving 40 trials in total.

\item \textbf{Instance-Level Generalization (OOD):} Evaluated on the Pick-and-Place task, this regime tests whether the policy can generalize to object instances that are unseen during training. We select 10 novel objects that are not included in the training dataset, and evaluate each object at the same 5 pre-designed tabletop positions. This produces 50 trials in total.

\item \textbf{Attribute Generalization (OOD):} Conducted on the Open-Steamer-Transfer-Bun task, this regime evaluates robustness to visual attribute shifts in both objects and surrounding scene. We construct 4 plate-variation settings by changing plate type, and 1 background-variation setting by changing the tablecloth color. Under each setting, we evaluate 8 different buns with variations in shape, color, and texture pattern. This yields 40 trials. 
\end{itemize}


\subsection{Training Details}
We find that embodiment-domain pure-video training of Cosmos is a critical initialization step before video-action joint optimization. In preliminary experiments, directly starting from the original Cosmos checkpoint and jointly training the video branch with the action head causes a sharp drop in downstream success rate. Therefore, we first train Cosmos with only the video denoising objective for approximately 4,000-5,000 steps on the corresponding dataset, and use the resulting checkpoint, rather than the original Cosmos weights, to initialize the video branch for video-action joint training. We use a total batch size of 256 and optimize the model with a cosine learning-rate schedule with a 5\% warm-up phase. All training runs are conducted on 32 GPUs, each with 140GB memory. The total training objective consists of the Cosmos denoising loss $\mathcal{L}_{\mathrm{vid}}$, the action denoising loss $\mathcal{L}_{\mathrm{act}}$, and the AGRA loss $\mathcal{L}_{\mathrm{AGRA}}$, with loss weights 1.0, 1.0, and 0.01, respectively.

For real-world experiments, after the initial pure-video adaptation stage, we train the full video-action model on the joint pretraining dataset for 60,000 steps. During this stage, the learning rate of Cosmos is set to $1\times10^{-5}$, while the learning rate of the action head is set to $1\times10^{-4}$. We then fine-tune the model on the task-specific fine-tuning datasets for 2,000 steps. During fine-tuning, Cosmos is frozen and only the action head is updated. We use an action horizon of $K=48$ for a single action chunk. Since Cosmos predicts 4 future latent frames, corresponding to 16 RGB frames, we sample the video frames every 3 frames when pairing video targets with actions. This makes the temporal span of the 16-frame video target consistent with the 48-step action chunk. 

For simulation experiments, the model is trained for 80,000 steps in the full-data regime, and trained for 40,000 steps in the few-shot regime. The learning rates of both Cosmos and the action head are set to $1\times10^{-4}$. We set the action chunk horizon to $K=16$. The corresponding video target contains 16 RGB frames, so no temporal interval sampling is required.

\section{Additional Experimental Results}

\subsection{Real-World Execution Cases}
We provide additional real-world execution cases of AGRA in Figure~\ref{fig:demo}. The examples cover both ID and OOD evaluation settings. In the OOD settings, AGRA is evaluated under semantic generalization, instance-level generalization, and attribute generalization. These cases show that AGRA can follow different language instructions, distinguish target containers or objects, and complete the instructed manipulation under variations in object instances, visual attributes, and scene appearance. Figure~\ref{fig:demo_compare} compares AGRA with the baseline WAM in real-world execution. Under the instructions ``put the ball in the box'' and ``put the peach in the box'', the baseline WAM fails to localize the target object accurately and produces spatially biased grasps. In contrast, AGRA consistently identifies the correct target region and executes more precise reaching and grasping motions.

\subsection{Analysis for Representation Alignment Layer}
We further analyze why the intermediate layer is a suitable location for applying representation alignment. Figure~\ref{fig:all_layer_text} visualizes the cross-attention between text tokens and video tokens inside Cosmos across different layers. For the instruction ``put the banana in the box,'' we visualize the attention maps corresponding to the tokens ``banana'' and ``box''. The results show that at the 8th layer, the attention maps are most spatially consistent with the corresponding objects in the generated video: the token ``banana'' attends to the banana region, and the token ``box'' attends to the box region. This phenomenon suggests that the 8th layer is where Cosmos exposes relatively clear instruction-conditioned semantic grounding. Applying AGRA at this stage is therefore reasonable: the model has already formed object-level semantic correspondence between language and visual regions, while deeper layers can still remain available for modeling motion, geometry, and fine-grained spatiotemporal dynamics. This supports our previous findings in Section~\ref{sec:exp_analysis} (a).

We also visualize the cross-attention maps in the action head, as shown in Figure~\ref{fig:all_layer_action}. This visualization reveals how the action decoder reads information from the predicted video representation when generating actions. At early action layers, the attention is relatively global and covers broad scene regions. This indicates that the action head first aggregates coarse contextual information, such as the scene layout, object distribution, and robot configuration. As the action prediction proceeds to middle and later layers, the attention becomes increasingly concentrated around the robot hands, especially the regions relevant to hand-object interaction. This progression suggests a hierarchical decoding process: the action head first establishes global task context and then focuses on local interaction regions needed for precise control.

\subsection{Visualization for Different Semantic Feature Types}
To complement the analysis in Section~\ref{sec:exp_analysis} (b), we visualize the feature spaces of different semantic visual encoders in Figure~\ref{fig:dino_feature}. Specifically, we compare DINOv2 and SigLIP features using PCA-based visualization. DINOv2 produces a more spatially structured and object-centric representation: object boundaries, geometric contours, and different functional regions are more clearly separated. In contrast, SigLIP features are more spatially diffuse.

\end{document}